\documentclass[10pt,twocolumn,letterpaper]{article}

\usepackage{iccv}
\usepackage{times}
\usepackage{epsfig}
\usepackage{graphicx}
\usepackage{amsmath}
\usepackage{amssymb}
\usepackage{booktabs}
\usepackage{diagbox}
\usepackage{pifont}
\usepackage{multirow}
\usepackage[abs]{overpic}
\usepackage{algorithm}
\usepackage[noend]{algpseudocode}
\usepackage{algorithmicx}
\usepackage{marvosym}
\usepackage{url}

\usepackage[breaklinks=true,bookmarks=false]{hyperref}

\DeclareMathOperator*{\argmin}{arg\,min}
\iccvfinalcopy 

\def\iccvPaperID{4167} 

\ificcvfinal\fi

\begin{document}

\title{Boosting Weakly Supervised Object Detection via Learning Bounding Box Adjusters}

\author{
   Bowen Dong\textsuperscript{\rm 1}\quad Zitong Huang\textsuperscript{\rm 1}\quad Yuelin Guo\textsuperscript{\rm 1}\quad Qilong Wang\textsuperscript{\rm 2}\quad Zhenxing Niu\textsuperscript{\rm 3}\quad  Wangmeng Zuo\textsuperscript{\rm 1,4\Letter} \\
   \textsuperscript{\rm 1}Harbin Institute of Technology \quad \textsuperscript{\rm 2}Tianjin University \quad \textsuperscript{\rm 3}Alibaba Group \quad \textsuperscript{\rm 4}Pazhou Lab, Guangzhou\\
   \small{\{cndongsky, zitonghuang99, zhenxingniu\}@gmail.com \quad gyl2565309278@163.com \quad qlwang@tju.edu.cn \quad wmzuo@hit.edu.cn}
}

\maketitle
\ificcvfinal\thispagestyle{empty}\fi

\begin{abstract}
   Weakly-supervised object detection (WSOD) has emerged as an inspiring recent topic to avoid expensive instance-level object annotations. 
   However, the bounding boxes of most existing WSOD methods are mainly determined by precomputed proposals, thereby being limited in precise object localization.
   In this paper, we defend the problem setting for improving localization performance by leveraging the bounding box regression knowledge from a well-annotated auxiliary dataset. 
   First, we use the well-annotated auxiliary dataset to explore a series of learnable bounding box adjusters (LBBAs) in a multi-stage training manner, which is class-agnostic. 
   Then, only LBBAs and a weakly-annotated dataset with non-overlapped classes are used for training LBBA-boosted WSOD. As such, our LBBAs are practically more convenient and economical to implement while avoiding the leakage of the auxiliary well-annotated dataset. 
   In particular, we formulate learning bounding box adjusters as a bi-level optimization problem and suggest an EM-like multi-stage training algorithm.
   Then, a multi-stage scheme is further presented for LBBA-boosted WSOD. 
   Additionally, a masking strategy is adopted to improve proposal classification. 
   Experimental results verify the effectiveness of our method. 
   Our method performs favorably against state-of-the-art WSOD methods and knowledge transfer model with similar problem setting. Code is publicly available at \url{https://github.com/DongSky/lbba_boosted_wsod}.
\end{abstract}

\vspace{-2em}
\section{Introduction}

Object detection \cite{girshick2014rcnn,girshickICCV15fastrcnn,renNIPS15fasterrcnn,Lin_2017_CVPR} has attracted considerable attention in computer vision community, and benefits a wide range of applications.
Along with the development of powerful convolutional neural networks (CNNs) and large-scale well-annotated datasets, the performance of object detection networks has achieved remarkable improvement.
Nevertheless, the success of object detection networks highly depends on precise but costly instance-level bounding box annotations of abundant images.
To alleviate this issue, weakly supervised object detection (WSOD) aiming at learning effective detection models with image-level supervision has emerged as an inspiring recent topic.

Existing WSOD methods~\cite{bilen2016weakly,tang2018pcl,Zeng_2019_ICCV,ren-wetectron2020} usually adopt the multiple instance learning (MIL) framework based on the precomputed proposals.
And most efforts have been given to improve proposal classification ability.
However, the bounding boxes of most existing methods are mainly determined by precomputed proposals, thereby being limited in precise object localization.
For single-phase WSOD methods~\cite{bilen2016weakly,tang2017multiple,tang2018pcl,Shen_2019_CVPR,li2019weakly}, the precomputed proposals classified to a specific class are directly taken as the detection results.
Bounding box regression branches are introduced in~\cite{yang2019towards,ren-wetectron2020,Zeng_2019_ICCV} and multi-phase training are adopted in~\cite{zhang2018w2f,Arun_2019}.
But they are usually supervised based on the pseudo ground-truths by selecting precomputed proposals with the highest scores.
In terms of localization performance, there remains a huge gap between WSOD methods and their fully-supervised counterparts.

Transfer learning has also been investigated to improve the localization performance of WSOD.
Lee \etal~\cite{ubbr2018} presented a universal bounding box regressor (UBBR) trained on a well-annotated auxiliary dataset for refining bounding boxes generated in WSOD.
Instead, Uijlings \etal~\cite{uijlings2018revisiting} trained a universal detector on the well-annotated source dataset, which is then transferred to WSOD as a generic proposal generator.
However, \cite{ubbr2018} and \cite{uijlings2018revisiting} adopt the single-stage transfer strategy, 
which actually are not specified to WSOD~\cite{bilen2016weakly,tang2017multiple,ubbr2018,uijlings2018revisiting} and suffer
from imperfect annotations in source domain \cite{lin2014microsoft,everingham2010pascal,uijlings2018revisiting}. 
Going beyond \cite{uijlings2018revisiting},  Zhong \etal~\cite{zhong2020boosting} trained and exploited the one-class universal detector (OCUD) in a progressive manner. 
In contrast, both the source well-annotated and target weakly annotated
datasets are required in the whole training process for OCUD~\cite{zhong2020boosting}. 
When the source dataset is private and is of large scale~\cite{sun2017revisiting,mahajan2018exploring}, it is preferred to avoid the direct joint use of the source and target datasets for WSOD with transfer learning. 
Instead, the owner of source datasets can first extract knowledge from data and then distribute knowledge instead of source datasets to the user for boosting WSOD.

In this paper, we follow the problem setting in~\cite{ubbr2018,uijlings2018revisiting}, and propose a learnable bounding box adjuster (LBBA) for boosting WSOD performance.
%
%
Specifically, we consider a well-annotated auxiliary dataset and a weakly annotated dataset.
Our method involves two subtasks, \ie, learning class-agnostic bounding box adjuster and training LBBA-boosted WSOD model.
In comparison to~\cite{ubbr2018,uijlings2018revisiting}, the LBBAs are specifically designed for improving WSOD performance by developing a multi-stage scheme.
Different from~\cite{zhong2020boosting}, only the LBBAs and weakly-annotated dataset are used for boosting WSOD, and thus our approach is practically convenient and economical for WSOD training while avoiding the leakage of the auxiliary dataset.

To better learn LBBAs from the well-annotated auxiliary dataset and exploit them to improve the performance of WSOD, we formulate the learning of LBBAs as a bi-level optimization problem and present an EM-like multi-stage training algorithm. 
In particular, the lower subproblem is formulated to learn a deep detection model by incorporating WSOD with LBBA-based regularization, while the upper subproblem is formulated to learn the boundary box adjuster for regressing the selected region proposals generated by WSOD towards the ground-truth bounding boxes.
With such formulation, the LBBAs can thus be learned for optimizing WSOD performance.
For solving the bi-level optimization problem, we adopt an EM-like multi-stage training algorithm by alternating between training LBBA and WSOD models.
Given the class-agnostic and multi-stage LBBAs, the training of LBBA-boosted WSOD also involves several stages.
In each stage, the final LBBA can be used to predict the bounding boxes based on the selected region proposals generated by WSOD, which are then used to train the WSOD models.

Nevertheless, our LBBAs improve localization performance but are limited in improving proposal classification. As a remedy, we introduce a masking strategy to improve the classification performance of the detector.
Specifically, a multi-label classifier is introduced to predict category confidence on image-level, which can further suppress scores of false-positive proposals of WSOD network.

Extensive experiments have been conducted to evaluate our proposed method.
Benefiting from the class-agnostic setting, LBBAs generalize well to new classes of objects and improves the localization performance of WSOD.
Our method performs favorably against state-of-the-art WSOD methods as well as knowledge transfer models with similar problem setting, \eg, UBBR~\cite{ubbr2018}. 
Contributions of this work can be summarized as follows:
\vspace{-0.6em}
\begin{itemize}
    \item [1)] Multi-stage learnable bounding box adjusters are presented for improving localization performance of WSOD, which is the core component of our proposed framework. Particularly, LBBAs make it feasible to use source and target datasets separately for training WSOD models, which is practically more convenient and economical.
    \vspace{-0.6em}
   \item [2)] A bi-level optimization formulation, as well as an EM-like multi-stage training algorithm, are suggested to learn LBBAs specified for optimizing WSOD. 
   \vspace{-0.6em}
   \item [3)] An effective masking strategy is introduced to improve the accuracy of the proposal classification branch.
   \vspace{-0.6em}
   \item [4)] Experimental results show our proposed method performs favorably against the state-of-the-art WSOD methods and knowledge transfer models with the similar problem setting.
   \vspace{-1em}
\end{itemize}

\begin{figure*}
\begin{center}
\includegraphics[width=6in]{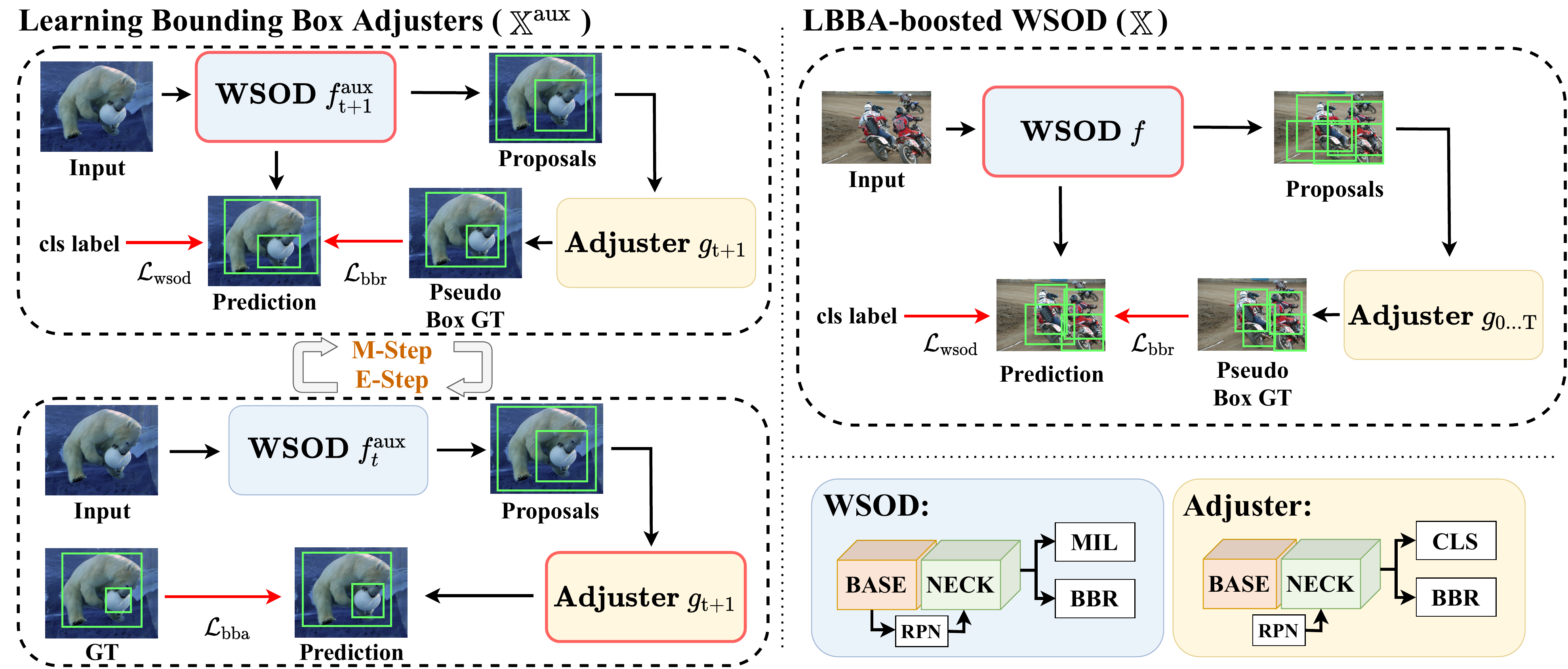}
\end{center}
\vspace{-1em}
   \caption{Illustration of our proposed method which includes two subtasks, \ie,  \textbf{learning bounding box adjusters} (left) and \textbf{LBBA-boosted WSOD} (right). 
   For learning bounding box adjusters, we adopt an EM-like algorithm.
   In \textbf{E-step}, adjuster $g$ predicts bounding boxes from proposals of $f^{\text{aux}}$ and supervised by ground-truths of $\mathbb{X}^{\text{aux}}$; In \textbf{M-step}, WSOD network $f^{\text{aux}}$ is supervised by image label as well as adjusted boxes from $g$ on $\mathbb{X}^{\text{aux}}$. For {LBBA-boosted WSOD}, WSOD network $f$ is supervised by image label and adjusted boxes from $g$ on $\mathbb{X}$. Finally, the learned ${f}$ is used for evaluation.}
\vspace{-1.5em}
\label{fig:pipeline}
\end{figure*}

\section{Related Work}
\subsection{Weakly Supervised Object Detection}
Weakly supervised object detection (WSOD) aims at training an effective detector only using image-level labels, and is usually formulated as a multiple instance learning (MIL) problem \cite{Dietterich1997SolvingTM}. Existing WSOD approaches can be roughly grouped into two categories: single-phase training methods and multi-phase training ones. For single-phase training methods, they rely on precomputed proposals \cite{uijlings2013selective,APBMM2014,zitnick2014edge} during training and testing. Specifically, Bilen \textit{et al.} \cite{bilen2016weakly} proposed a two-stream detection network (WSDDN) as the basic proposal classifier. To improve \emph{proposal classification ability}, OICR \cite{tang2017multiple} and PCL \cite{tang2018pcl} proposed online classifier refinement module. OIM \cite{lin2020object} proposed spatial and appearance graphs with object instance reweighted loss to resolve part domination. 
SDCN \cite{li2019weakly} and WS-JDS \cite{Shen_2019_CVPR} introduced segmentation branch and collaboration loop to reweight proposals. 
As for improving \emph{proposal localization ability}, Yang \textit{et al.} \cite{yang2019towards}, WSOD2 \cite{Zeng_2019_ICCV} and MIST \cite{ren-wetectron2020} introduced bounding box regression into WSOD network, where proposals with highest scores are selected as pseudo ground-truths to supervise bounding box regression branch. 

For multi-phase training methods~\cite{zhang2018w2f,zigzag2018, li2019weakly,Wan_2019, Gao_2019_ICCV}, an additional detector is further trained by selecting proposals with the highest scores as pseudo ground-truths based on the output of trained WSOD network in the prior phase  \cite{girshickICCV15fastrcnn}. Any single-phase methods \cite{tang2017multiple,tang2018pcl,yang2019towards,Arun_2019} can be extended to multi-phase setting by this procedure. Current multi-phase training methods focus on how to select pseudo ground-truths with the highest scores. 
%
%
However, these approaches rely on only selected precomputed proposals to localize objects or supervise box regression branch, low precision proposals restrict the localization ability of WSOD approaches. Different from the above methods, we aim at resolving this issue by using learnable bounding box adjusters, which provide more precise pseudo boxes supervision to help WSOD network obtain better object localization ability.

\subsection{Transfer Learning in WSOD}
Transfer learning based WSOD usually leverages an auxiliary dataset to provide semantic information or class-agnostic information to help WSOD networks train on weakly-annotated target dataset. 
Previous works~\cite{Guillaumin2012LargescaleKT,hoffman2014lsda,tang2016large} focused on \emph{transferring semantic information} between strong classifier and weakly supervised detector. Among them, Hoffman \textit{et al.} \cite{hoffman2014lsda} proposed LSDA, which introduces category specific adaptation to adapt a classifier into target detection dataset. Tang \textit{et al.} \cite{tang2016large} further extended LSDA by building visual similarity and semantic relatedness. 
Nonetheless, above methods are not proposed for improving bounding box regression. 

Recently, several approaches \cite{wslat2015,msd2018,uijlings2018revisiting,ubbr2018,zhong2020boosting} have been studied to exploit transfer learning for \emph{improving object localization performance}.
\cite{wslat2015,msd2018,uijlings2018revisiting,zhong2020boosting} proposed to learn proposal generators to help WSOD network locate novel objects on weakly-annotated target dataset. 
Among them, \cite{wslat2015,msd2018,uijlings2018revisiting} trained proposal generators merely using the auxiliary dataset, while Zhong \textit{et al.} trained generator on both auxiliary dataset and weakly-annotated dataset progressively to generalize better on target dataset. 
Instead, Lee \textit{et al.} \cite{ubbr2018} proposed a box refinement module, which takes the random transformations of ground-truth boxes as the input to learn class-agnostic box regressor, and also exhibits certain generalization ability on target weakly-annotated dataset. However, the real boxes generated during WSOD training may be quite different from those by random transformations, making the learned regressor not tailored to WSOD.
In comparison to existing methods, our LBBAs can be considered as the multi-stage training of box refinement modules only using the auxiliary dataset, and achieves very competitive box regression performance on weakly-annotated dataset. Different from UBBR\cite{ubbr2018}, our method dynamically takes the proposals generated by WSOD as the input to train LBBA, and thus is expected to achieve improved detection performance.

\section{Proposed Method}
\subsection{Problem Setting and Notations}
In this work, we follow the problem setting in~\cite{wslat2015,msd2018,uijlings2018revisiting,ubbr2018} for WSOD by using a well-annotated auxiliary dataset $\mathbb{X}^{\text{aux}}$ and a weakly annotated dataset $\mathbb{X}$.
In particular, $\mathbb{X}^{\text{aux}}$ is first used to train class-agnostic learnable bounding box adjusters (LBBAs).
Then, we utilize both LBBAs and any weakly annotated dataset $\mathbb{X}$ to learn a better WSOD model. 
For the image-level weakly annotated dataset $\mathbb{X} = \{\mathbf{I}, \mathbb{P},\mathbf{y}\}$, $\mathbf{I}$ denotes an image from $\mathbb{X}$, and $\mathbf{y}$ denotes the corresponding image-level labels. 
For the end of WSOD, MCG~\cite{APBMM2014} and selective search~\cite{uijlings2013selective} are used to extract a set of precomputed proposals $\mathbb{P} = \{\mathbf{p}\}$ for each image $\mathbf{I}$.
Besides $\mathbb{X}$, we also introduce a well-annotated auxiliary dataset $\mathbb{X}^{\text{aux}} = \{(\mathbf{I}^{\text{aux}},\mathbb{P}^{\text{aux}}, \{\mathbf{b}^{\text{aux}}\}, \mathbf{y}^{\text{aux}})\}$. 
For an image $\mathbf{I}^{\text{aux}}$ from $\mathbb{X}^{\text{aux}}$, $\mathbf{y}^{\text{aux}}$ denotes the image-level labels, and $\{\mathbf{b}^{\text{aux}}\}$ denotes the annotated bounding boxes. 
To aid WSOD, we also give the precomputed proposals $\mathbb{P}^{\text{aux}} = \{\mathbf{p}^{\text{aux}}\}$ of $\mathbf{I}^{\text{aux}}$. 
To show the generalization ability of LBBA, we assume the object classes in $\mathbb{X}$ are not overlapped with those in $\mathbb{X}^{\text{aux}}$.

We argue that the above problem setting is both practically valuable and convenient in implementation. 
Albeit weakly-supervised learning is preferred for object detection, several well-annotated datasets, \eg, COCO~\cite{lin2014microsoft}, have already been publicly available.    
Our problem setting allows the learned bounding box adjusters to be deployed in training new classes of object detector, thereby being expected to be advantageous to conventional WSOD solely relying on $\mathbb{X}$.  
In OCUD~\cite{zhong2020boosting}, the well-annotated dataset $\mathbb{X}^{\text{aux}}$ is directly incorporated with the weakly-annotated dataset $\mathbb{X}$ for WSOD. 
In our problem setting, the well-annotated dataset $\mathbb{X}^{\text{aux}}$ can be safely abandoned after learning bounding box adjusters. 
Then, LBBAs can be incorporated with any weakly annotated dataset $\mathbb{X}$ for WSOD. 
We note that LBBAs can avoid the direct leakage of well-annotated dataset $\mathbb{X}^{\text{aux}}$ to the users with weakly annotated dataset $\mathbb{X}$, thereby being more convenient, economic, and secure in practice.

\subsection{Overview}
In general, our method involves two subtasks, \ie, (i) learning bounding box adjusters, and (ii) LBBA-boosted WSOD.  
The overall training procedure is shown in Fig. \ref{fig:pipeline}.
To better draw the LBBAs from well-annotated auxiliary dataset, we formulate the learning of bounding box adjusters as a bi-level optimization problem.
In the lower-subproblem, we use a WSOD method and current LBBA $g_{{t}}$ to update the object detection model $f_{{t+1}}$ from  $\{(\mathbf{I}^{\text{aux}},\mathbb{P}^{\text{aux}},  \mathbf{y}^{\text{aux}})\}$.
So the learned $f_{{t+1}}$ can also be represented as a function of LBBA. 
Therefore, the upper-subproblem is formulated to learn $g_{{t+1}}$ specified for optimizing the performance of the weakly-supervised object detector by using the well-annotated data $\{(\mathbf{I}^{\text{aux}},\{\mathbf{b}^{\text{aux}}\},  \mathbf{y}^{\text{aux}})\}$.
In each stage, we first update the learning of bounding box adjuster $g_{{t+1}}$ by fixing $f_{{t}}$, and then update the weakly-supervised object detector $f_{{t+1}}$ by fixing LBBA $g_{{t+1}}$.   
With several stages (${T}=3$) of training. We can obtain a set of LBBA models \{$g_{{0}}, ..., g_{{T}}\}$ with one for each stage.

For LBBA-boosted WSOD, the well-annotated dataset $\mathbb{X}^{\text{aux}}$ can be abandoned, and only the LBBA models \{$g_{{0}}, ..., g_{{T}}\}$ and the weakly annotated dataset $\mathbb{X}$ are required.   
LBBA-boosted WSOD also involves several stages (\ie, ${T}$). 
In each stage (\eg, ${t}$), we  use the current object detector $f_{{t}}$ to obtain a set of selected proposals and exploit the stage-wise LBBA $g_{{t}}$ for bounding box adjustment. 
Then, the adjusted bounding boxes are introduced into the WSOD model for updating $f_{{t+1}}$. 
In the following, after introducing the baseline WSOD model used in this work, we present our solutions to the subtasks of both learning bounding box adjusters and LBBA-boosted WSOD in detail.

\subsection{Baseline WSOD Model}
%
%
To learn both bounding box regression and proposal classification from weakly-annotated dataset, we adopt the method proposed in \cite{ijcai2018-135,yang2019towards} as our baseline network $f(\mathbf{I},\mathbb{P};\theta_{{f}})$. 
Here, $\theta_{{f}}$ denotes the model parameters of the object detector. 
Specifically, the network $f(\mathbf{I},\mathbb{P};\theta_{{f}})$ involves a basic multi-instance-learning  (MIL) branch as well as an independent bounding box regression (BBR) branch. %
Given an input image $\mathbf{I}$ with image-level label $\mathbf{y}=\left\{\mathbf{y}_{{1}},...,\mathbf{y}_{{C}}\right\}$ as well as $R$ precomputed proposals $\mathbb{P}_{\text{mil}}=\left\{\mathbf{p}_{\text{mil},{1}},...,\mathbf{p}_{\text{mil},{R}}\right\}$, MIL branch generates two $R \times C$ logits $\mathbf{x}^{\text{cls}}$ and $\mathbf{x}^{\text{det}}$, which are passed through softmax layers.
Then, a fusion score $\textbf{s} = \sigma_{\text{cls}}(\mathbf{x}^{\text{cls}}) \cdot \sigma_{\text{det}}(\mathbf{x}^{\text{det}})$ can be computed by performing element-wise product on those of classification and localization. 
Finally, the image-level score of class $c$ can be attained by
\vspace{-0.5em}
\begin{equation}
   \label{eqn:fusion_to_image}
\textbf{q}_{{c}}=\sum\nolimits_{i=1}^{{R}} \textbf{s}_{{i,c}}. 
\vspace{-0.5em}
\end{equation}
And the MIL branch can be optimized by 
\vspace{-0.5em}
\begin{equation}
   \mathcal{L}_{\text{wsddn}} = \text{BCE}(\mathbf{q}, \mathbf{y};{\theta}_{\text{f}}),
   \vspace{-0.5em}
\end{equation}
where $\text{BCE}(\cdot, \cdot)$ denotes the binary cross-entropy loss. 
To improve detection quality, we also introduce pseudo label mining strategy and construct instance refinement branch optimized by a set of weighted instance refinement loss $\mathcal{L}_{\text{r}}$ \cite{tang2017multiple,tang2018pcl,ren-wetectron2020}. 

In typical single phase WSOD, the precomputed proposals classified to a specific class are taken as the detection results. 
To improve the object localization performance, we follow~\cite{ijcai2018-135} to introduce an RPN module into our WSOD network for generating region proposals $\mathbb{P}_{\text{rpn}} = \{\mathbf{p}_{\text{rpn}}\}$.
%
%
%
%
Then, all proposals from $ \mathbb{P}  = \mathbb{P}_{\text{mil}} \cup \mathbb{P}_{\text{rpn}} $ are sent into bounding box regression branch to generate corresponding localization outputs.
Following standard Faster R-CNN~\cite{renNIPS15fasterrcnn}, both RPN module and bounding box regression branch are trained by the losses $\mathcal{L}_{\text{rpn-cls}}$, $\mathcal{L}_{\text{rpn-det}}$ and $\mathcal{L}_{\text{det}}$ defined on pseudo ground-truth instances selected by refinement scores. 
Thus, the learning objective of our baseline WSOD model can be written as,
\begin{equation}
   \mathcal{L}_{\text{wsod}}=\mathcal{L}_{\text{wsddn}}+\mathcal{L}_{\text{r}}+\mathcal{L}_{\text{rpn-cls}}+\mathcal{L}_{\text{rpn-det}}+\mathcal{L}_{\text{det}},
\end{equation}
where $\mathcal{L}_{\text{r}}$ and $\mathcal{L}_{\text{rpn-cls}}$ are the cross-entropy losses supervised by pseudo class labels on the selected proposals, while $\mathcal{L}_{\text{rpn-det}}$ and $\mathcal{L}_{\text{det}}$ are the smooth-L1 losses~\cite{girshickICCV15fastrcnn} supervised by the proposal boxes of pseudo ground-truths. Note that we follow the same strategy of OICR~\cite{tang2017multiple} to generate pseudo ground-truths.

We note that the bounding box regression branch in baseline WSOD model is learned based on the supervision from the precomputed proposals, which naturally are not precise enough. 
In the subsequent subsections, we learn a set of bounding box adjusters to provide better ground-truth for supervising the bounding box regression branch, thereby being beneficial to detection performance.   
Moreover, we use the above baseline WSOD model as an example to show the effectiveness of the learned bounding box adjusters.
Actually, our proposed method is independent with most existing WSOD methods and can be incorporated with them to further boost detection performance. 
And we will illustrate this point in the experiments. 


\begin{algorithm}
   \caption{Learning Bounding Box Adjusters}
   \label{alg:lbba}
   \begin{algorithmic}[1]
      \Require Auxiliary dataset $\mathbb{X}^{\text{aux}}$, adjuster network $g$ , WSOD network $f^{\text{aux}}$, stage num ${T}$
      \Ensure Adjuster parameters $\{{\theta}_{g}^{{0}}\dots{\theta}_{g}^{{T}}\}$
      \State Initialize ${\theta}_{g}^{{0}}$ on $\mathbb{X}^{\text{aux}}$
      \State $\theta^{{0}}_{f^{\text{aux}}}\leftarrow \argmin\limits_{\theta_{f^{\text{aux}}}} \mathcal{L}_{\text{wsod}}+\mathcal{L}_{\text{bbr}}$
      \For{${t}=0...{T-1}$}
      \State \textbf{E-Step:}
         \State $\theta^{{t+1}}_{\text{g}} \leftarrow \argmin\limits_{\theta_{g}} \mathcal{L}_{\text{bba}}$
      \State \textbf{M-Step:}
         \State $\theta^{{t+1}}_{f^{\text{aux}}}\leftarrow \argmin\limits_{\theta_{f^{\text{aux}}}} \mathcal{L}_{\text{wsod}}+\mathcal{L}_{\text{bbr}}$
      \EndFor
      \State $\textbf{return}$ $\{{\theta}_{g}^{{0}}\dots{\theta}_{g}^{{T}}\}$
   \end{algorithmic}
\end{algorithm}

\subsection{Learning Bounding Box Adjusters}
\subsubsection{Bi-level Optimization Formulation}
To formulate our weakly supervised object detection problem elegantly, we first revisit the traditional EM algorithm for weakly supervised learning. 
In particular, E-step is used to update latent variable $\hat{\text{b}}$,
\begin{equation}\label{eq:pre_e}
   \vspace{-0.5em}
   \hat{\text{b}} = \arg\max\limits_{\text{b}_{\text{latent}}}\log P(\mathbf{y}|\text{b}_{\text{latent}}) - \mathcal{L}(\text{b}_{\text{latent}},f(\mathbf{I},\mathbb{P};\theta_{{f}})).
   \vspace{-0.5em}
\end{equation} 
For WSOD with box regression, $\mathbf{y}$ is image class labels, $\mathcal{L}$ is defined as box regression loss (\textit{e.g.}, smooth L1 loss~\cite{girshickICCV15fastrcnn} for bounding box regression), $\hat{\text{b}}$ means latent bounding box variables, and $P(\mathbf{y}|\text{b}_{\text{latent}})$ is probability of $\mathbf{y}$ with given $\text{b}_{\text{latent}}$ in WSOD training. And $f(\mathbf{I},\mathbb{P};\theta_{{f}})$ is bounding box output from WSOD network $f$ with corresponding parameters $\theta_{{f}}$. We mainly discuss $\mathcal{L}$ in next paragraphs.
Then, M-step is deployed to update the model parameters $\theta_{{f}}$.
\begin{equation}
   \vspace{-0.5em}
   \theta_{{f}} = \arg\min\limits_{\theta_{f}}\mathcal{L}(\hat{\text{b}},f(\mathbf{I},\mathbb{P};\theta_{{f}})),
   \vspace{-0.5em}
\end{equation}
where $\mathcal{L}$ is a combination of weakly supervised object detection loss $\mathcal{L}_{\text{wsod}}$ and bounding box regression loss $\mathcal{L}_{\text{bbr}}$.

As mentioned above, previous methods utilize precomputed proposals as well as pseudo ground-truth mining in E-step, and then update box regression branch of WSOD network in M-step.
However, optimizing $P(\mathbf{y}|\text{b}_{\text{latent}})$ in E-step with only image-level supervision to improve quality of $\hat{\text{b}}$ is difficult.
Besides, when optimizing $\mathcal{L}$ in E-step, precomputed proposals are designed for generating region proposals for box regression of object detection, which are not suitable for final object localization. 
To tackle this problem, we want to use extra well-annotated data to supervise a learnable model, make it generate more precise $\hat{\text{b}}$ in E-step.
Therefore, we first introduce a full-annotated auxiliary dataset $\mathbb{X}^{\text{aux}}$ to provide class-agnostic localization supervision. 
And then, we aim to introduce a class-agnostic Learnable Bounding Box Adjuster (LBBA) $g(\mathbf{I}^{\text{aux}}, \mathbb{P}^{\text{aux}}; \theta_g)$ trained on $\mathbb{X}^{\text{aux}}$, which takes the selected proposals from $ \mathbb{P}^{\text{aux}}  = \mathbb{P}^{\text{aux}}_{\text{mil}} \cup \mathbb{P}^{\text{aux}}_{\text{rpn}} $ as the input. 
For each $\mathbf{p}^{\text{aux}} \in \mathbb{P}^{\text{aux}}$, $g(\mathbf{I}^{\text{aux}}, \mathbb{P}^{\text{aux}}; \theta_g)$ aims to predict a more precise estimation of bounding box $\hat{\mathbf{b}}^{\text{aux}}$, which is then used to supervise the bounding box regression branch in WSOD.   
Denoted by $\tilde{\mathbf{b}}^{\text{aux}}$ the output of bounding box regression.
We apply smooth L1 loss~\cite{girshickICCV15fastrcnn} $\mathcal{L}_{\text{bbr}}$ for supervising bounding box regression branch of $f$,
\begin{equation}
   \vspace{-0.5em}
    \mathcal{L}_{\text{bbr}} = \sum\nolimits_{\mathbf{p}^{\text{aux}} \in \mathbb{P}^{\text{aux}}}
   {\text{Smooth}}_{L1}(\hat{\mathbf{b}}^{\text{aux}}, \tilde{\mathbf{b}}^{\text{aux}};{\theta}_{{f}}).
   \vspace{-0.5em}
 \end{equation}
Using the ground-truth bounding box $\mathbf{b}^{\text{aux}}$ from $\mathbb{X}^{\text{aux}}$, we further introduce a loss $\mathcal{L}_{\text{bba}}$ for supervising the learning of bounding box adjusters,
\begin{equation}
   \vspace{-0.5em}
    \mathcal{L}_{\text{bba}} = \sum\nolimits_{\mathbf{p}^{\text{aux}} \in \mathbb{P}^{\text{aux}}}
   {\text{Smooth}}_{L1}({\mathbf{b}}^{\text{aux}}, \tilde{\mathbf{b}}^{\text{aux}};{\theta}_{{g}}).
   \vspace{-0.5em}
 \end{equation}
To this end, we suggest to utilize LBBA $g$ to generate latent variable $\hat{\text{b}}_{\text{aux}}$ on $\mathbb{X}^{\text{aux}}$.

\vspace{-0.5em}
\begin{equation}
   \vspace{-1em}
   \begin{split}
   &\hat{\text{b}}_{\text{aux}} = g(\mathbf{I}^{\text{aux}}, \mathbb{P}^{\text{aux}}; \theta_g) \\
   \theta_g = \arg&\min\limits_{\theta_g}\mathcal{L}_{\text{bba}}(\{\text{b}^{\text{aux}}\},g(\mathbf{I}^{\text{aux}}, \mathbb{P}^{\text{aux}}; \theta_g))
   \end{split}
   \vspace{-0.7em}
\end{equation}

After introducing LBBA $g$ into WSOD, our WSOD problem can be transferred into a \textbf{bi-level optimization problem}, here we state how to build bi-level optimization. 

\noindent\textbf{Lower subproblem.} During M-step, WSOD network $f$ is supervised by both image class label $\mathbf{y}$ as well as latent variable $\hat{\text{b}}^{\text{aux}}$, which is output of LBBA network $g(\mathbf{I}^{\text{aux}}, \mathbb{P}^{\text{aux}}; \theta_g)$. Therefore we update parameters of WSOD network $\theta_{f^{\text{aux}}}$ by minimizing $\mathcal{L}_{\text{wsod}}+\mathcal{L}_{\text{bbr}}$, which is shown as follows,
\begin{equation}\label{eq:m_lbba}
   \hspace{-4mm}\theta_{f^{\text{aux}}}\text{=}\arg\min\limits_{\theta_{f^{\text{aux}}}} (\mathcal{L}_{\text{wsod}}\text{+}\mathcal{L}_{\text{bbr}}) (\hat{\text{b}}^{\text{aux}},f^{\text{aux}}(\mathbf{I}^{\text{aux}},\mathbb{P}^{\text{aux}};\theta_{f^{\text{aux}}}))
   \vspace{-0.5em}
\end{equation}

\noindent\textbf{Upper subproblem.} Taking above equations into consideration, WSOD parameters $\theta_{f^{\text{aux}}}$ can be seen as a function of LBBA parameters $\theta_g$ (\textit{i.e.}, $\theta_{f^{\text{aux}}}(\theta_g)$). Thus, in E-step the upper subproblem on $\theta_g$ is defined for optimizing $\mathcal{L}_{\text{bba}}$ on the WSOD network $f^{\text{aux}}(\mathbf{I}^{\text{aux}},\mathbb{P}^{\text{aux}};\theta_{f^{\text{aux}}}(\theta_g))$, 
\begin{equation}\label{eq:e_lbba}
   \theta_g\text{=}\arg\min\limits_{\theta_g} \mathcal{L}_{\text{bba}}(\{\text{b}^{\text{aux}}\},f^{\text{aux}}(\mathbf{I}^{\text{aux}},\mathbb{P}^{\text{aux}};\theta_{f^{\text{aux}}}(\theta_g)))
   \vspace{-0.5em}
\end{equation}
where $g$ generates adjusted bounding box regression for given proposals from WSOD $f^{\text{aux}}$. Thus upper subproblem has transferred into a fully-supervised setting.

\subsubsection{EM-like Multi-stage Training Algorithm}
From Eqns.~(\ref{eq:m_lbba},\ref{eq:e_lbba}), the direct optimization of $\theta_g$ involves the cumbersome computation of the partial gradient $(\partial{\mathcal{L}_{\text{bbr}}}/\partial{\theta_f})(\partial{\theta_f}/\theta_g)$.  %
Briefly, direct joint training of two networks to solve this bi-level optimization problem is harmful to the generalization ability of LBBA. And EM-like training strategy can keep that of LBBA.
Therefore, to avoid this issue, we suggest an EM-like multi-stage training algorithm. 
Suppose that $f_{{t}}(\mathbf{I}^{\text{aux}}, \mathbb{P}^{\text{aux}}_{\text{mil}}; \theta_f^{t})$ and $g_{{t}}(\mathbf{I}^{\text{aux}}, \mathbb{P}^{\text{aux}}; \theta_g^{{t}})$ are the learned models at stage ${t}$. 
In the E-step, we use $f_{{t}}(\mathbf{I}^{\text{aux}}, \mathbb{P}^{\text{aux}}_{\text{mil}}; \theta_f^{{t}})$ to generate and select the proposals $\mathbb{P}^{\text{aux}}$, which are then deployed to learn $g_{{t+1}}(\mathbf{I}^{\text{aux}}, \mathbb{P}^{\text{aux}}; \theta_g^{{t+1}})$. 
In the M-step, we use $\theta_g^{{t+1}}$ to substitute $\theta_g$ in $\mathcal{L}_{\text{bbr}}$, and obtain $f_{{t+1}}(\mathbf{I}^{\text{aux}}, \mathbb{P}^{\text{aux}}_{\text{mil}}; \theta_f^{{t+1}})$ by solving the lower subproblem, thereby resulting in our EM-like multi-stage training algorithm.
In the following, we explain the initialization, E-step, and M-step in more detail.

\textbf{Initialization.} To begin with, we utilize $\mathbb{X}^{\text{aux}}$ to train a two-stage detector with class-agnostic bounding box regression branch, which is then used as the bounding box adjuster $g_{{0}}$ at stage $t = 0$. 
Then, the selected proposals from $\mathbb{P}^{\text{aux}}_{\text{mil}}$ are fed into $g_{{0}}$ to generate the adjusted bounding boxes for supervising the learning of WSOD model $f_{{0}}$. 

\textbf{E-step.} Given the learned model parameters $\theta_f^{{t}}$ of $f_{{t}}$ at stage ${t}$, the E-step aims at learning the bounding box adjuster $g_{{t+1}}$ with the model parameters $\theta_g^{{t+1}}$. 
For an image $\mathbf{I}^{\text{aux}}$ from $\mathbb{X}^{\text{aux}}$, we utilize the RPN module of $f_{{t}}$ to generate a set of region proposals $\mathbb{P}_{\text{rpn}}^{\text{aux}}$. 
We empirically find that it is better to take the region proposal instead of the bounding box predicted by $f_{{t}}$ as the input to $g_{{t+1}}$. 
Moreover, both the precomputed and the generated proposals $\mathbb{P}_{\text{mil}}^{\text{aux}} \cup \mathbb{P}_{\text{rpn}}^{\text{aux}}$ are beneficial to the training of $g_{{t+1}}$.  
Thus, we use $f_{t}$ with the parameters $\theta_f^{{t}}$ to predict the bounding boxes, and decode them to generate the corresponding selected proposals $\mathbb{P}^{\text{aux}}_{\text{wsod}}$ from $\mathbb{P}_{\text{mil}}^{\text{aux}} \cup \mathbb{P}_{\text{rpn}}^{\text{aux}}$. 
The model $g_{{t+1}}$ takes $\mathbb{P}^{\text{aux}}_{\text{wsod}}$ as the input to predict a set of adjusted bounding boxes $\{ \hat{\mathbf{b}}^{\text{aux}} \}$.
With the ground-truth bounding boxes from $\mathbb{X}^{\text{aux}}$, we  train the bounding box adjuster $g_{{t+1}}$ with the parameters $\theta_g^{{t+1}}$ at stage ${t+1}$ by minimizing the loss ${\mathcal{L}_{\text{bba}}}$.

\textbf{M-step.} With the help of the learned model parameters $\theta_g^{{t+1}}$ of $g_{{t+1}}$, the M-step learns the WSOD model $f_{{t+1}}$ with the model parameters $\theta_f^{{t+1}}$. 
In the forward propagation,  an image $\mathbf{I}^{\text{aux}}$ from $\mathbb{X}^{\text{aux}}$ is fed into the current WSOD model to generate a number of region proposals $\mathbb{P}_{\text{rpn}}^{\text{aux}}$ and bounding boxes.  
Then, we decode the predicted bounding boxes to obtain the selected proposals $\mathbb{P}^{\text{aux}}_{\text{wsod}}$ from $\mathbb{P}_{\text{mil}}^{\text{aux}} \cup \mathbb{P}_{\text{rpn}}^{\text{aux}}$. 
Taking $\mathbb{P}^{\text{aux}}_{\text{wsod}}$ as the input, the adjusted bounding boxes predicted by the LBBA $g_{{t+1}}$ are then used to define the loss $\mathcal{L}_{\text{bbr}}$.
Finally, the WSOD model $f_{{t+1}}$ with the model parameters $\theta_f^{{t+1}}$ can be trained by minimizing the combined loss $\mathcal{L}_{\text{wsod}} + \mathcal{L}_{\text{bbr}}$.

To sum up, after the initialization, our training algorithm alternates between the E-step and M-step for $T$ times. 
Hence, it is a multi-stage training scheme, where we run the E-step and M-step once in each stage. 
The training process of LBBA is given in Algorithm \ref{alg:lbba}.
%


\subsection{LBBA-boosted WSOD}
\label{sec:mkd}
After learning bounding box adjusters, the well-annotated auxiliary dataset can be abandoned. 
For the LBBA-boosted WSOD task, we only require a weakly-annotated dataset $\mathbb{X}$ as well as a set of learned bounding box adjusters $\{g_{{0}}, ..., g_{{T}}\}$.
The multi-stage scheme is also adopted to train WSOD, and we use stage ${t}$ as an example to illustrate the training process. 
In particular, an image $\mathbf{I}$ from $\mathbb{X}$ is fed into the current WSOD model to generate a number of region proposals $\mathbb{P}_{\text{rpn}}$ and bounding boxes.  
Then, we decode the predicted bounding boxes to obtain the selected proposals $\mathbb{P}_{\text{wsod}}$ from $\mathbb{P}_{\text{mil}} \cup \mathbb{P}_{\text{rpn}}$. 
Taking $\mathbb{P}_{\text{wsod}}$ as the input, the adjusted bounding boxes predicted by the LBBA $g_{{t}}$ are then used to define the loss $\mathcal{L}_{\text{bbr}}$.
%
%
Finally, the WSOD model $f_{{t}}$ with the model parameters $\theta_f^{{t}}$ can be trained by minimizing the combined loss $\mathcal{L}_{\text{wsod}} + \mathcal{L}_{\text{bbr}}$. 
After ${T}$ stages of training, the WSOD model at stage ${T}$, \ie, $f_{{T}}$ with parameters $\theta_f^{{T}}$, can be kept and applied to the test images. The training process of LBBA-boosted WSOD is given in Algorithm \ref{alg:lbba-wsod}.

Nonetheless, we empirically find that updating WSOD network with only the last $g_{T}$ can attain a similar performance. Hence we can build a lighter pipeline by only using the last $g_{T}$.

\begin{figure*}
\begin{center}
\includegraphics[width=6in]{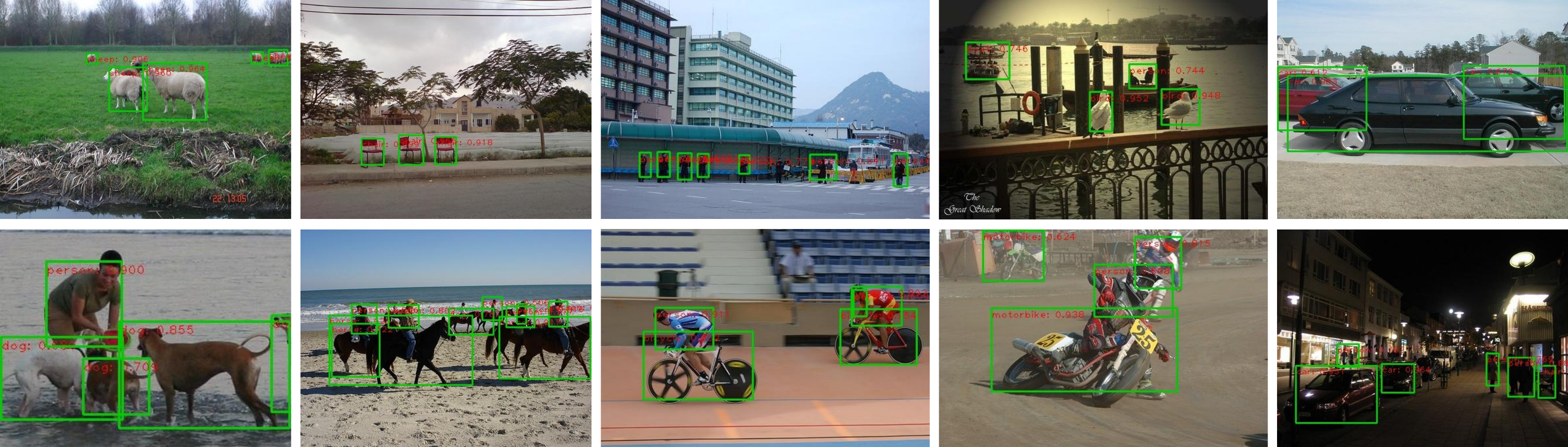}
\end{center}
\vspace{-1em}
    \caption{Visualization results of our method on PASCAL VOC 2007, which has the ability to generate precise bounding boxes.}
\vspace{-1em}
\label{fig:vis-2007}
\end{figure*}   
\begin{algorithm}
   \caption{LBBA-boosted WSOD}
   \label{alg:lbba-wsod}
   \begin{algorithmic}[1]
      \Require Weakly-annotated dataset $\mathbb{X}$, stage num ${T}$, adjuster network $g$, adjuster parameters $\{{\theta}_{g}^{{0}}\dots{\theta}_{g}^{{T}}\}$, WSOD network $f$
      \Ensure WSOD network parameters ${\theta}_{{f}}^{{T}}$
      \For{${t}=0...{T}$}
      
      \State 
      $\theta_{g} \leftarrow \theta_{g}^{t}$
      \State $\theta^{{t}}_{f}\leftarrow \argmin\limits_{\theta_{f}} \mathcal{L}_{\text{wsod}}+\mathcal{L}_{\text{bbr}}$
      \EndFor
      \State $\textbf{return}$ ${\theta}_{{f}}^{{T}}$
   \end{algorithmic}
\end{algorithm}
\vspace{-1em}

\subsection{Masking Strategy for Proposal Classification}
\label{sec:MLIC}
The above training algorithm can improve localization ability of WSOD network but cannot improve the ability of proposal classification.
To further improve the detection performance, we introduce an additional multi-label image classifier $h(\mathbf{I};\theta_{h})$ and present a classification score masking strategy. 
During training, we utilize images and corresponding image labels of dataset $\mathbb{X}$ to train $h$; during testing, given input image $\mathbf{I}$, we obtain image classification score by $\hat{{\textbf{s}}} = h(\mathbf{I};\theta_{h})$,
where $\hat{{\textbf{s}}} \in \mathbb{R}^{1 \times C}$ is per-class prediction scores of $\mathbf{I}$.
Therefore, we can judge which categories should not be included in $\mathbf{I}$, and suppress the corresponding output of WSOD. 
Specifically, we select a threshold $\tau$ (\ie, = -3.0), if $\hat{\text{s}}_{c}<\tau$, we assert that the category $c$ is not appeared in this image. 
Therefore, for each category c with $\hat{s}_{c} < \tau$,  score of $i$-th proposal $\mathbf{\hat{b}}_{i,c}$ is set to 0 to eliminate wrong predictions.

\section{Experiments}
\label{sec:exp}
\subsection{Datasets and Evaluation Metrics}
\textbf{Auxiliary Dataset.}
MS-COCO 2017~\cite{lin2014microsoft} is a large-scale object detection dataset. Note that MS-COCO dataset includes 80 different object classes. To eliminate semantic overlap and show the generalization ability of our method, we construct a subset of MS-COCO by excluding PASCAL VOC classes instance annotations and call it COCO-60. As such, COCO-60 dataset contains $\sim$98K training images and $\sim$4K validation images, respectively.

\textbf{Target Datasets.}
PASCAL VOC 2007 and 2012 datasets contain 9,963 images and 22,531 images collected from 20 object classes. For fair comparison, we use \textit{trainval} set for training WSOD networks and report evaluation results on \textit{test} set. During the training process, only image-level labels are used as supervision. We also utilized other datasets to evaluate our LBBA, see the suppl. for details.

\textbf{Evaluation Metrics.}
Since our method aims at improving object detection performance, Average Precision (AP) is used as the basic evaluation metric in our experiments. We also adopt CorLoc \cite{corloc2012} as another evaluation metric.

\begin{table}[t]
    \caption{Single model detection results on PASCAL VOC 2007 and 2012, where $\text{~}^{+}$ means the results with multi-scale testing, $\text{~}^{*}$ means joint training of WSOD models on auxiliary dataset and weakly-annotated dataset.}
    \centering
    \footnotesize{
    \begin{tabular}{l | c | c}
    \specialrule{.15em}{.05em}{.05em}
    Methods & ~~mAP~~ (07) & ~~mAP~~ (12) \\
    \hline
    $\text{OICR}^{+}$ \cite{tang2017multiple} & 41.2 & 37.9 \\
    $\text{PCL}^{+}$ \cite{tang2018pcl} & 43.5 & 40.6\\
    $\text{Yang }\textit{et al.}^{+}$ \cite{yang2019towards} & 51.5 & 46.8 \\
    $\text{WSOD 2}^{+}$ \cite{Zeng_2019_ICCV} & 53.6 & 47.2\\
    Arun \textit{et al.} \cite{Arun_2019} & 52.9 & 48.4 \\
    $\text{C-MIDN}^{+}$ \cite{Gao_2019_ICCV} & 52.6 & 50.2 \\
    $\text{MIST (Full)}^{+}$ \cite{ren-wetectron2020} & {54.9} & {52.1}   \\
    \hline
    $\text{MSD-Ens}^{+}$ \cite{msd2018} & 51.1 & - \\
    OICR+UBBR \cite{ubbr2018} & 52.0 & - \\
    \hline
    $\text{Zhong \textit{et al.} (R50-C4)}^{*}$ \cite{zhong2020boosting} & {\textbf{57.8}} & - \\
    $\text{Zhong \textit{et al.} (R50-C4)}^{+*}$ \cite{zhong2020boosting} & {\textbf{59.7}} & - \\
    \hline
    \textbf{Ours} & \textbf{56.5} & \textbf{54.7}   \\
    $\textbf{Ours}^{+}$ & \textbf{56.6} & \textbf{55.4}   \\
    \specialrule{.15em}{.05em}{.05em}
    \textbf{Upper bounds:} \\
    \hline
    Faster R-CNN \cite{renNIPS15fasterrcnn} & {69.9} & 67.0\\
    \specialrule{.15em}{.05em}{.05em}
    \end{tabular}
    }
    \label{table:map-0712}
    \vspace{-2em}
    \end{table}

\begin{table}[t]
    \caption{Single model correct localization (CorLoc) results on PASCAL VOC 2007 and 2012, where $\text{~}^{+}$ means the results with multi-scale testing, $\text{~}^{*}$ means joint training of WSOD models on auxiliary dataset and weakly-annotated dataset.}
    \centering
    \footnotesize{
    \begin{tabular}{l | c | c}
    \specialrule{.15em}{.05em}{.05em}
    Methods & ~~CorLoc~~ (07) & ~~CorLoc~~ (12) \\
    \hline
    $\text{OICR}^{+}$ \cite{tang2017multiple} & 60.6 & 62.1 \\
    $\text{PCL}^{+}$ \cite{tang2018pcl} & 62.7 & 63.2 \\
    $\text{Li}^{+}$ \cite{li2019weakly} & 68.6 & 67.9 \\
    $\text{Yang }\textit{et al.}^{+}$ \cite{yang2019towards} & 68.0 & 69.5 \\
    $\text{WSOD 2}^{+}$ \cite{Zeng_2019_ICCV} & 69.5 & 71.9\\
    Arun~ \textit{et al.}\cite{Arun_2019} & 70.9 & 69.5 \\
    $\text{C-MIL}^{+}$ \cite{Wan_2019} & 65.0 & 67.4 \\
    $\text{MIST (Full)}^{+}$ \cite{ren-wetectron2020} & {68.8} & {70.9}   \\
    \hline
    WSLAT-Ens \cite{wslat2015} & 58.8 & - \\
    $\text{MSD-Ens}^{+}$ \cite{msd2018} & 66.8 & - \\
    OICR+UBBR \cite{ubbr2018} & 47.6 & - \\
    \hline
    $\text{Zhong \textit{et al.} (R50-C4)}^{*}$ \cite{zhong2020boosting} & {\textbf{73.6}} & - \\
    $\text{Zhong \textit{et al.} (R50-C4)}^{+*}$ \cite{zhong2020boosting} & \textbf{74.4} & - \\
    \hline
    \textbf{Ours} & \textbf{72.3} & \textbf{73.2}   \\
    $\textbf{Ours}^{+}$ & \textbf{72.5} & \textbf{73.7} \\
    \specialrule{.15em}{.05em}{.05em}
    \end{tabular}
    }
    \label{table:corloc-0712}
    \vspace{-2em}
    \end{table}

\subsection{Comparison with State-of-the-arts}
We \emph{state the implementation details in the suppl.} and we build up all experiments based on it. 
We compare our method with several state-of-the-art WSOD approaches in terms of detection and localization performance on PASCAL VOC datasets. As suggested in \cite{bilen2016weakly,tang2017multiple,tang2018pcl,yang2019towards,ren-wetectron2020,Arun_2019,zhong2020boosting}, we report detection results on \textit{test} set and localization results on \textit{trainval} set, respectively. 
Table~\ref{table:map-0712} compares the results of different state-of-the-art WSOD approaches on PASCAL VOC 2007 and 2012 datasets. It can be seen that our LBBA improves OICR and OICR+REG over 15.3\% and 5.0\% on PASCAL VOC 2007 dataset, respectively. Furthermore,  our method performs better than all competing methods, except Zhong \textit{et al.}~\cite{zhong2020boosting}. Note that \cite{zhong2020boosting} uses stronger backbone model and knowledge transfer strategy by directly incorporating source and target datasets. Moreover, the auxiliary dataset adopted in Zhong \etal is different from ours (See the suppl. for more details). As shown in Fig.~\ref{fig:vis-2007}, our method has the ability to generate precise bounding boxes. On PASCAL VOC 2012, our LBBA is superior to all competing methods and obtains more than 1\% gains over all WSOD approaches. Experimental results show that our method is effective in improving the detection performance of WSOD.

We further evaluate the localization performance of our method. Table~\ref{table:corloc-0712} lists results of several state-of-the-art WSOD approaches on PASCAL VOC 2007 and 2012. Our LBBA outperforms OICR by 11.7\% and also improves the baseline OICR+REG over 4.3\% on PASCAL VOC 2007 dataset. Besides, our LBBA performs better than all competing methods. Meanwhile, on PASCAL VOC 2012, our LBBA is also superior to all competing methods, and obtains 1.3\% gain over WSOD 2\cite{Zeng_2019_ICCV}.
In comparison to Zhong \etal \cite{zhong2020boosting}, our LBBA-based method employs a weaker backbone model and avoids the direct joint use of the source and target datasets, while still achieving competitive CorLoc results under the settings of both single-scale testing and multi-scale testing. 
The above results show that our LBBA-based method is effective in improving the localization performance of WSOD.

\subsection{Ablation Study}
Additionally, we employ PASCAL VOC 2007 to assess the effect of some key components on our LBBA. We state a more detailed ablation study in the suppl..

\textbf{Backbone Models of Adjuster $g$.} In this work, Faster R-CNN~\cite{renNIPS15fasterrcnn}  is used as adjuster. Here, we first evaluate the effect of backbone models on adjuster $g$. To this end, we compare two CNN architectures as backbone models of Faster R-CNN, i.e., ResNet-50~\cite{He2016DeepRL} and VGG-16~\cite{simonyan2014very}. Particularly, we set iterations $T$ of multi-stage learning to $3$ and adopt WSDDN~\cite{bilen2016weakly} as WSOD network $f$. The compared results on VOC 07 are listed in Table~\ref{table:voc-coco-res}, from which we can see that adjuster $g$ with backbone of ResNet-50 outperforms one with backbone of VGG-16 by 2.5\% and 2.6\% in terms of mAP and CorLoc, respectively. These results show that our method can benefit from a stronger adjuster, which encourages us to develop more effective adjusters. 

\begin{table}[t]
	\caption{Comparison of different backbone models of Adjuster $g$ on VOC 07, where iterations $T$ of multi-stage learning is set to $3$ while WSDDN~\cite{bilen2016weakly} is used as WSOD network $f$.}
	\centering
	\footnotesize{
		\begin{tabular}{l|cc}
			\specialrule{.15em}{.05em}{.05em}
			 Backbone of Adjuster $g$ & mAP (VOC 07) & CorLoc (VOC 07) \\
			\hline
			 VGG-16 & {50.2} & {67.7}  \\
			\hline
			 R50-C4 & \textbf{52.7} & \textbf{70.3}  \\
			\specialrule{.15em}{.05em}{.05em}
		\end{tabular}
	}
	\vspace{-1em}
	\label{table:voc-coco-res}
\end{table}

\begin{table}[t]
	\caption{Comparison of various WSOD networks $f$ on VOC 07.}
	\centering
	\footnotesize{
		\begin{tabular}{l |  c c }
			\specialrule{.15em}{.05em}{.05em}
			Method & mAP (VOC 07)  & CorLoc (VOC 07) \\
			\hline
			Baseline (WSDDN) & {46.6} & {64.7}  \\
			Baseline (OICR) & {48.6} & {66.8}  \\
			Baseline (OICR+\cite{ren-wetectron2020}) & {51.4} & {64.9}  \\
			\hline
			Ours (WSDDN) & \textbf{52.7} & \textbf{70.3}  \\
			Ours (OICR) & \textbf{55.1} & \textbf{71.0}  \\
			Ours (OICR+\cite{ren-wetectron2020}) & \textbf{55.8} & \textbf{71.6}  \\
			\specialrule{.15em}{.05em}{.05em}
		\end{tabular}
	}
	\vspace{-1em}
	\label{table:voc-coco-head}
	\vspace{-1em}
\end{table} 

\textbf{Effect of WSOD network $f$.} After determining backbone model of adjuster $g$, we access the impact of WSOD network $f$. Specifically, we consider three methods (i.e., WSDDN+REG \cite{bilen2016weakly}, OICR+REG \cite{tang2017multiple} and OICR+REG with top p\% pseudo label mining \cite{ren-wetectron2020}) for our WSOD network $f$, and compare our LBBA with the original methods (i.e., baseline). The iterations $T$ of multi-stage learning is set to $3$, and the results of different WSOD networks $f$ are given in Table \ref{table:voc-coco-head}. First, our LBBA achieves clear performance gains (more than 4\%) over the baseline methods for all choices of WSOD networks in terms of mAP and CorLoc. It demonstrates that the proposed LBBA methods can be well generalized to various WSOD networks. Second, our LBBA benefits from stronger WSOD networks, and so we compare with state-of-the-arts by using OICR+\cite{ren-wetectron2020} as WSOD network $f$. 

\begin{table}[t]
   \caption{Results of adjuster $g$ and WSOD network $f$ on COCO-60 and VOC 07 using different learning strategies, respectively}
       \centering
           \footnotesize{
       \begin{tabular}{l | c | c }
       \specialrule{.15em}{.05em}{.05em}
       Learning Strategy & Adjuster mAP (COCO-60) & mAP (VOC 07)  \\
       \hline
       $T$=0 & 29.1 & {53.1}   \\
       $T$=1 & 29.6 & 54.9 \\
       $T$=2 & 29.9 & {55.7} \\
       $T$=3 & \textbf{30.9} & \textbf{55.8} \\
       \hline
       LBBA-MCG & 29.6 & 54.3 \\
       \specialrule{.15em}{.05em}{.05em}
       \end{tabular}
       }
       \vspace{-2.2em}
       \label{table:table-k}
   \end{table}

\textbf{Multi-stage LBBAs.} The proposed multi-stage learning strategy of LBBAs involves two core factors, \ie, number of iterations ($T$) and learnable, auxiliary WSOD network $f^{\text{aux}}$. By fixing  WSOD network $f$ and adjuster $g$ respectively be OICR+\cite{ren-wetectron2020} and Faster R-CNN with backbone of ResNet-50, we assess the effects of number of iterations ($T$) and $f^{\text{aux}}$ on our LBBA method. To this end, we learn bounding box adjusters by setting $T$ from 0 to 3. Besides, we replace learnable $f^{\text{aux}}$ by using MCG to generate proposals, namely LBBA-MCG. Table \ref{table:table-k} gives the results of adjuster $g$ and WSOD network $f$ on COCO-60 and VOC 07 using different learning strategies, respectively. It can be seen that increasing iterations ($T$) can improve performance of both adjuster $g$ and WSOD network $f$. However, performance of WSOD network $f$ is sightly improved,  when number of iterations $T > 2$. Therefore, $T=3$ is a good choice to balance efficiency and effectiveness. These results clearly demonstrate the effectiveness of our multi-stage learning strategy. The learnable $f^{\text{aux}}$ with 3 iterations is superior to LBBA-MCG by 1.3\% and 1.5\% for adjuster $g$ and WSOD network $f$, showing the significance of learnable $f^{\text{aux}}$.

\section{Conclusion} 
In this paper, we presented a knowledge transfer based WSOD method. 
Our proposed method involves two subtasks, \ie, learning bounding box adjusters and LBBA-boosted WSOD. 
For the former subtask, we suggested a bi-level optimization formulation on the auxiliary dataset and an EM-like training algorithm to learn multi-stage and class-agnostic LBBAs specified for optimizing WSOD performance. 
For the later subtask, we adopted a multi-stage scheme to utilize only the LBBAs and weakly-annotated dataset for WSOD. 
Additionally, a masking strategy is adopted to improve proposal classification for benefiting detection performance.
Experimental results show that our proposed method performs favorably against the state-of-the-art WSOD methods and knowledge transfer model with similar problem setting~\cite{ubbr2018,msd2018,wslat2015,zhong2020boosting}.
Nonetheless, we mainly focus on transferring across classes in this paper, while the transferring across domains is not specifically considered. In the future, we will explore suitable domain generalization methods for coping with this issue.

\section*{Acknowledgement}
This work was supported in part by the National Natural Science Foundation of China under grant No.s U19A2073 and 61806140, and Natural Science Foundation of Tianjin under grant No. 20JCQNJC1530.
{\small
\bibliographystyle{ieee_fullname}
\bibliography{egbib}
}
\clearpage
\appendix
\section{Discussion of EM-like training algorithm}
The reason why EM-like training is necessary is that the problem is formulated as a bi-level optimization problem, direct joint training to solve this problem is harmful to the generalization ability of LBBA. And EM-like training can keep that of LBBA. Here we state why formulating WSOD problem as bi-level optimization.

In particular, E-step is used to update latent variable $\hat{\text{b}}$,
\begin{equation}\label{eq:pre_e}
   \hat{\text{b}} = \arg\max\limits_{\text{b}_{\text{latent}}}\log P(\mathbf{y}|\text{b}_{\text{latent}}) - \mathcal{L}(\text{b}_{\text{latent}},f(\mathbf{I},\mathbb{P};\theta_{{f}})).
   \vspace{-0.5em}
\end{equation} 
For WSOD with box regression, $\mathbf{y}$ is image class labels, $\mathcal{L}$ is defined as box regression loss (\textit{e.g.}, smooth L1 loss~\cite{girshickICCV15fastrcnn} for bounding box regression), $\hat{\text{b}}$ means latent bounding box variables, and $P(\mathbf{y}|\text{b}_{\text{latent}})$ is probability of $\mathbf{y}$ with given $\text{b}_{\text{latent}}$ in WSOD training. And $f(\mathbf{I},\mathbb{P};\theta_{{f}})$ is bounding box output from WSOD network $f$ with corresponding parameters $\theta_{{f}}$. We mainly discuss $\mathcal{L}$ in next paragraphs.
Then, M-step is deployed to update the model parameters $\theta_{{f}}$.
\begin{equation}
   \theta_{{f}} = \arg\min\limits_{\theta_{f}}\mathcal{L}(\hat{\text{b}},f(\mathbf{I},\mathbb{P};\theta_{{f}})),
   \vspace{-0.5em}
\end{equation}
where $\mathcal{L}$ is a combination of weakly supervised object detection loss $\mathcal{L}_{\text{wsod}}$ and bounding box regression loss $\mathcal{L}_{\text{bbr}}$.

As mentioned above, previous methods utilize precomputed proposals as well as pseudo ground-truth mining in E-step, and then update box regression branch of WSOD network in M-step.
However, optimizing $P(\mathbf{y}|\text{b}_{\text{latent}})$ in E-step with only image-level supervision to imporve quality of $\hat{\text{b}}$ is difficult.
Besides, when optimizing $\mathcal{L}$ in E-step, precomputed proposals are designed for generating region proposals for box regression of object detection, which are not suitable for final object localization. 
To tackle this problem, we want to use extra well-annotated data to supervise a learnable model, make it generate more precise $\hat{\text{b}}$ in E-step.
%
Therefore, we aim to introduce a class-agnostic Learnable Bounding Box Adjuster (LBBA) $g(\mathbf{I}^{\text{aux}}, \mathbb{P}^{\text{aux}}; \theta_g)$ trained on a full-annotated auxiliary dataset $\mathbb{X}^{\text{aux}}$. To this end, we suggest to utilize LBBA $g$ to generate latent variable $\hat{\text{b}}^{\text{aux}}$ on $\mathbb{X}^{\text{aux}}$.

\begin{equation}
   \begin{split}
   &\hat{\text{b}}_{\text{aux}} = g(\mathbf{I}^{\text{aux}}, \mathbb{P}^{\text{aux}}; \theta_g) \\
   \theta_g = \arg&\min\limits_{\theta_g}\mathcal{L}_{\text{bba}}(\{\text{b}^{\text{aux}}\},g(\mathbf{I}^{\text{aux}}, \mathbb{P}^{\text{aux}}; \theta_g))
   \end{split}
   \vspace{-0.5em}
\end{equation}
After introducing LBBA $g$ into WSOD, our WSOD problem can be transferred into a \textbf{bi-level optimization problem}, here we state how to build bi-level optimization. 

\noindent\textbf{Lower subproblem.} During M-step, WSOD network $f$ is supervised by both image class label $\mathbf{y}$ as well as latent variable $\hat{\text{b}}^{\text{aux}}$, which is output of LBBA network $g(\mathbf{I}^{\text{aux}}, \mathbb{P}^{\text{aux}}; \theta_g)$. Therefore we update parameters of WSOD network $\theta_{f^{\text{aux}}}$ by minimizing $\mathcal{L}_{\text{wsod}}+\mathcal{L}_{\text{bbr}}$, which is shown as Eq.~\ref{eq:m_lbba}. And Eq.~\ref{eq:m_lbba} also stands for the lower subproblem of bi-level optimization.
\begin{equation}\label{eq:m_lbba}
   \hspace{-4mm}\theta_{f^{\text{aux}}}\text{=}\arg\min\limits_{\theta_{f^{\text{aux}}}} (\mathcal{L}_{\text{wsod}}\text{+}\mathcal{L}_{\text{bbr}}) (\hat{\text{b}}^{\text{aux}},f^{\text{aux}}(\mathbf{I}^{\text{aux}},\mathbb{P}^{\text{aux}};\theta_{f^{\text{aux}}}))
   \vspace{-0.5em}
\end{equation}

\noindent\textbf{Upper subproblem.} Thus, taking above equations into consideration, WSOD parameters $\theta_{f^{\text{aux}}}$ can be seen as a function of LBBA parameters $\theta_g$ (\textit{i.e.}, $\theta_{f^{\text{aux}}}(\theta_g)$). Thus, in E-step the upper subproblem on $\theta_g$ is defined for optimizing $\mathcal{L}_{\text{bba}}$ on the WSOD network $f^{\text{aux}}(\mathbf{I}^{\text{aux}},\mathbb{P}^{\text{aux}};\theta_{f^{\text{aux}}}(\theta_g))$, 
\begin{equation}\label{eq:e_lbba}
   \theta_g\text{=}\arg\min\limits_{\theta_g} \mathcal{L}_{\text{bba}}(\{\text{b}^{\text{aux}}\},f^{\text{aux}}(\mathbf{I}^{\text{aux}},\mathbb{P}^{\text{aux}};\theta_{f^{\text{aux}}}(\theta_g)))
   \vspace{-0.5em}
\end{equation}
where $g$ generates adjusted bounding box regression for given proposals from WSOD $f^{\text{aux}}$. Thus upper subproblem has transferred into a fully-supervised setting. Furthermore, to ease the training difficulty of the upper subproblem and improve the precision of $\hat{\text{b}}^{\text{aux}}$, we modify the upper subproblem by requiring LBBA accurately predicts the ground-truth boxes, resulting in the following bi-level optimization formulation.
\begin{equation}\label{eqn:bilevel}
   \begin{split} 
      \hspace{-4mm}&\min\limits_{{\theta}_{\text{g}}} \mathcal{L}_{\text{bba}}(\{\mathbf{b}^{\text{aux}}\},g(\mathbf{I}^{\text{aux}}, f^{\text{aux}}(\mathbf{I}^{\text{aux}},\mathbb{P}^{\text{aux}};\theta_{f^{\text{aux}}}); \theta_g)) \\
   \hspace{-4mm}&\textit{ s.t.}{\theta}_{\text{f}} \text{=} \arg\min\limits_{{\theta}_{\text{f}}} \mathcal{L}_{\text{wsod}}\text{+}\mathcal{L}_{\text{bbr}}(\hat{\text{b}}^{\text{aux}},f^{\text{aux}}(\mathbf{I}^{\text{aux}},\mathbb{P}^{\text{aux}};\theta_{f^{\text{aux}}})).
\end{split}
\end{equation}

\begin{table*}[t]
    \caption{Single model detection per-class results on PASCAL VOC 2007, where $\text{~}^{+}$ means the results with multi-scale testing, $\text{~}^{*}$ means joint training of WSOD models on the auxiliary dataset and weakly-annotated dataset.}
    \centering
    \resizebox{\textwidth}{!}{
    {
    \begin{tabular}{l | c c c c c c c c c c c c c c c c c c c c | c}
    \specialrule{.15em}{.05em}{.05em}
    Methods & Aero & Bike & Bird & Boat & Bottle & Bus & Car & Cat & Chair & Cow & Table & Dog & Horse & Motor & Person & Plant & Sheep & Sofa & Train & ~~TV~~ & ~~AP~~ \\
    \hline
    WSDDN~\cite{bilen2016weakly} &39.4 & 50.1 & 31.5 & 16.3 & 12.6 & 64.5 & 42.8 & 42.6 & 10.1 & 35.7 & 24.9 & 38.2 & 34.4 & 55.6 & 9.4 & 14.7 & 30.2 & 40.7 & 54.7 & 46.9 & 34.8 \\
    $\text{OICR}^{+}$ \cite{tang2017multiple} & 58.0 & 62.4 & 31.1 & 19.4 & 13.0 & 65.1 & 62.2 & 28.4 & 24.8 & 44.7 & 30.6 & 25.3 & 37.8 & 65.5 & 15.7 & 24.1 & 41.7 & 46.9 & 64.3 & 62.6 & 41.2 \\
    $\text{PCL}^{+}$ \cite{tang2018pcl} & 54.4 & 69.0 & 39.3 & 19.2 & 15.7 & 62.9 & 64.4 & 30.0 & 25.1 & 52.5 & 44.4 & 19.6 & 39.3 & 67.7 & 17.8 & 22.9 & 46.6 & 57.5 & 58.6 & 63.0 & 43.5 \\
    $\text{Yang }\textit{et al.}^{+}$ \cite{yang2019towards} & 57.6 & 70.8 & 50.7 & 28.3 & 27.2 & 72.5 & 69.1 & 65.0 & 26.9 & 64.5 & 47.4 & 47.7  & 53.5 & 66.9 & 13.7 & 29.3 & 56.0 & 54.9 & 63.4 & 65.2 & 51.5 \\
    $\text{C-MIDN}^{+}$ \cite{Gao_2019_ICCV} & 53.3 & 71.5 & 49.8 & 26.1 & 20.3 & 70.3 & 69.9 & 68.3 & 28.7 & 65.3 & 45.1 & {64.6} & 58.0 & 71.2 & 20.0 & 27.5 & 54.9 & 54.9 & {69.4} & 63.5 & 52.6 \\
    Arun \textit{et al.} \cite{Arun_2019} & 66.7 & 69.5 & 52.8 & 31.4 & 24.7 & {74.5} & 74.1 & 67.3 & 14.6 & 53.0 & 46.1 & 52.9 & 69.9 & 70.8 & 18.5 & 28.4 & 54.6 & {60.7} & 67.1 & 60.4 & 52.9 \\
    $\text{WSOD2}^{+}$ \cite{Zeng_2019_ICCV} & 65.1 & 64.8 & {57.2} & {39.2} & 24.3 & 69.8 & 66.2 & 61.0 & 29.8 & 64.6 & 42.5 & 60.1 & {71.2} & 70.7 & 21.9 & 28.1 & 58.6 & 59.7 & 52.2 & 64.8 & 53.6 \\
    MIST-Full \cite{ren-wetectron2020} & {68.8} & {77.7} & 57.0 & 27.7 & {28.9} & 69.1 & {74.5} & 67.0 & {32.1} & {73.2} & 48.1 & 45.2 & 54.4 & {73.7} & {35.0} & 29.3 & {64.1} & 53.8 & 65.3 & {65.2} & {54.9}  \\
    \hline
    $\text{MSD-Ens}^{+}$ \cite{msd2018} & 70.5 & 69.2 & 53.3 & 43.7 & 25.4 & 68.9 & 68.7 & 56.9 & 18.4 & 64.2 & 15.3 & 72.0 & 74.4 & 65.2 & 15.4 & 25.1 & 53.6 & 54.4 & 45.6 & 61.4 & 51.1 \\
    OICR+UBBR \cite{ubbr2018} & 59.7 & 44.8 & 54.0 & 36.1 & 29.3 & 72.1 & 67.4 & 70.7 & 23.5 & 63.8 & 31.5 & 61.5 & 63.7 & 61.9 & 37.9 & 15.4 & 55.1 & 57.4 & 69.9 & 63.6 & 52.0 \\
    \hline
    \textbf{Ours} & 65.4 & 73.7 & 53.1 & 44.8 & 27.3 & 73.1 & {73.7} & 72.2 & 29.8 & 69.2 & 51.1 & 68.7 & 56.4 & 71.8 & 20.3 & 27.1 & 61.4 & 60.3 & 65.5 & {65.9} & \textbf{56.5}  \\
    $\textbf{Ours}^{+}$ & 70.3 & 72.3 & 48.7 & 38.7 & 30.4 & 74.3 & 76.6 & 69.1 & 33.4 & 68.2 & 50.5 & 67.0 & 49.0 & 73.6 & 24.5 & 27.4 & 63.1 & 58.9 & 66.0 & 69.2 & \textbf{56.6} \\
    \specialrule{.15em}{.05em}{.05em}
    \textbf{Upper bounds:} \\
    \hline
    Faster R-CNN \cite{renNIPS15fasterrcnn} & 70.0 &  80.6  & 70.1 &  57.3  & 49.9  & 78.2  & 80.4  & 82.0  &  52.2 &  75.3  & 67.2  & 80.3  & 79.8  & 75.0  & 76.3  & 39.1  & 68.3 &  67.3  & 81.1 &  67.6  & {69.9} \\
    $\text{Zhong \textit{et al.} (R50-C4)}^{*}$ \cite{zhong2020boosting} & 64.4 & 45.0 & 62.1 & 42.8 & 42.4 & 73.1 & 73.2 & 76.0 & 28.2 & 78.6 & 28.5 & 75.1 & 74.6 & 67.7 & 57.5 & 11.6 & 65.6 & 55.4 & 72.2 & {61.3} & {57.8} \\
    $\text{Zhong \textit{et al.} (R50-C4)}^{+*}$ \cite{zhong2020boosting} & 64.8 & 50.7 & 65.5 & 45.3 & 46.4 & 75.7 & 74.0 & 80.1 & 31.3 & 77.0 & 26.2 & 79.3 & 74.8 & 66.5 & 57.9 & 11.5 & 68.2 & 59.0 & 74.7 & 65.5 & 59.7 \\
    \specialrule{.15em}{.05em}{.05em}
    \end{tabular}
    }}
    \label{table:per-cls-voc07}
    \vspace{1em}
\end{table*}

\begin{table*}[t]
    \caption{Single model detection results on PASCAL VOC 2012, where $\text{~}^{+}$ means the results with multi-scale testing, $\text{~}^{*}$ means joint training of WSOD models on the auxiliary dataset and weakly-annotated dataset.}
    \centering
    \resizebox{\textwidth}{!}{
    {
    \begin{tabular}{l | c c c c c c c c c c c c c c c c c c c c | c}
    \specialrule{.15em}{.05em}{.05em}
    Methods & Aero & Bike & Bird & Boat & Bottle & Bus & Car & Cat & Chair & Cow & Table & Dog & Horse & Motor & Person & Plant & Sheep & Sofa & Train & ~~TV~~ & ~~AP~~ \\
    \hline
    $\text{OICR}^{+}$ & 67.7 & 61.2 & 41.5 & 25.6 & 22.2 & 54.6 & 49.7 & 25.4 & 19.9 & 47.0 & 18.1 & 26.0 & 38.9 & 67.7 & 2.0 & 22.6 & 41.1 & 34.3 & 37.9 & 55.3 & 37.9 \\
    $\text{PCL}^{+}$ \cite{tang2018pcl} & 58.2 & 66.0 & 41.8 & 24.8 & 27.2 & 55.7 & 55.2 & 28.5 & 16.6 & 51.0 & 17.5 & 28.6 & 49.7 & 70.5 & 7.1 & 25.7 & 47.5 & 36.6 & 44.1 & 59.2 & 40.6\\
    $\text{Yang }\textit{et al.}^{+}$ & 64.7 & 66.3 & 46.8 & 28.5 & 28.4 & 59.8 & 58.6 & 70.9 & 13.8 & 55.0 & 15.7 & 60.5 &  {63.9} & 69.2 & 8.7 & 23.8 & 44.7 &  {52.7} & 41.5 & 62.6 & 46.8 \\
    $\text{WSOD2}^{+}$ \cite{Zeng_2019_ICCV} &-&-&-&-&-&-&-&-&-&-&-&-&-&-&-&-&-&-&-&-& 47.2\\
    Arun \textit{et al.} \cite{Arun_2019} &-&-&-&-&-&-&-&-&-&-&-&-&-&-&-&-&-&-&-&-& 48.4 \\
    $\text{C-MIDN}^{+}$ \cite{Gao_2019_ICCV} & 72.9 & 68.9 & 53.9 & 25.3 & 29.7 & 60.9 & 56.0 &  {78.3} & 23.0 & 57.8 & 25.7 &  {73.0} & 63.5 & 73.7 & 13.1 & 28.7 & 51.5 & 35.0 & 56.1 & 57.5 & 50.2 \\
    $\text{MIST (Full)}^{+}$ \cite{ren-wetectron2020}  & {78.3} & {73.9} & {56.5} & {30.4} & {37.4} & {64.2} & {59.3} & 60.3 & {26.6} & {66.8} & 25.0 & 55.0 & 61.8 & {79.3} & 14.5 & {30.3} & {61.5} & 40.7 & {56.4} & {63.5} & {52.1}   \\
    \hline
    \textbf{Ours}  & {77.0} & {71.0} & {62.0} & {40.0} & {37.5} & {67.4} & {62.5} & 68.3 & {23.6} & {71.4} & 25.6 & 78.4 & 71.9 & {74.3} & 6.7 & {29.2} & {62.8} & 50.6 & {47.8} & {62.1} & \textbf{54.5}   \\
    $\textbf{Ours}^{+}$ & 78.6 & 71.5 & 62.7 & 41.3 & 38.6 & 68.8 & 64.1 & 71.0 & 23.2 & 70.5 & 24.2 & 79.1 & 74.1 & 75.3 & 6.5 & 29.7 & 63.4 & 51.8 & 50.2 & 63.9 & \textbf{55.4} \\
    \specialrule{.15em}{.05em}{.05em}
    \textbf{Upper bounds:} \\
    \hline
    Faster R-CNN \cite{renNIPS15fasterrcnn} & 82.3 & 76.4 & 71.0 & 48.4 & 45.2 & 72.1 & 72.3 & 87.3 & 42.2 & 73.7 & 50.0 & 86.8 & 78.7 & 78.4 & 77.4 & 34.5 & 70.1 & 57.1 & 77.1 & 58.9 & 67.0\\
    \specialrule{.15em}{.05em}{.05em}
    \end{tabular}
    }}
    \label{table:per-cls-voc12}
    \end{table*}

\begin{table*}[tbh]
    \caption{Single model correct localization (CorLoc) results on PASCAL VOC 2007, where $\text{~}^{+}$ means the results with multi-scale testing, $\text{~}^{*}$ means joint training of WSOD models on the auxiliary dataset and weakly-annotated dataset.}
    \centering
    \resizebox{\textwidth}{!}{
    {
    \begin{tabular}{l | c c c c c c c c c c c c c c c c c c c c | c}
    \specialrule{.15em}{.05em}{.05em}
    Methods & Aero & Bike & Bird & Boat & Bottle & Bus & Car & Cat & Chair & Cow & Table & Dog & Horse & Motor & Person & Plant & Sheep & Sofa & Train & ~~TV~~ & CorLoc \\
    \hline
    WSDDN~\cite{bilen2016weakly} & 65.1 & 58.8 & 58.5 & 33.1 & 39.8 & 68.3 & 60.2 & 59.6 & 34.8 & 64.5 & 30.5 & 43.0 & 56.8 & 82.4 & 25.5 & 41.6 & 61.5 & 55.9 & 65.9 & 63.7 & 53.5 \\
    $\text{OICR}^{+}$ & 81.7 & 80.4 & 48.7 & 49.5 & 32.8 & 81.7 & 85.4 & 40.1 & 40.6 & 79.5 & 35.7 & 33.7 & 60.5 & 88.8 & 21.8 & 57.9 & 76.3 & 59.9 & 75.3 & {81.4} & 60.6 \\
    $\text{PCL}^{+}$ \cite{tang2018pcl} & 79.6 & 85.5 & 62.2 & 47.9 & 37.0 & 83.8 & 83.4 & 43.0 & 38.3 & 80.1 & 50.6 & 30.9 & 57.8 & 90.8 & 27.0 & 58.2 & 75.3 & {68.5} & 75.7 & 78.9 & 62.7 \\
    $\text{Li}^{+}$ \cite{li2019weakly} & 85.0 & 83.9 & 58.9 & 59.6 & 43.1 & 79.7 & 85.2 & 77.9 & 31.3 & 78.1 & 50.6 & 75.6 & 76.2 & 88.4 & 49.7 & 56.4 & 73.2 & 62.6 & 77.2 & 79.9 & 68.6 \\
    $\text{C-MIL}^{+}$ \cite{Wan_2019} &-&-&-&-&-&-&-&-&-&-&-&-&-&-&-&-&-&-&-&-& 65.0 \\
    $\text{Yang }\textit{et al.}^{+}$ & 80.0 & 83.9 & 74.2 & 53.2 & 48.5 & 82.7 & 86.2 & 69.5 & 39.3 & 82.9 & 53.6 & 61.4& 72.4 & 91.2 & 22.4 & 57.5 & {83.5} & 64.8 & 75.7 & 77.1 & 68.0 \\
    $\text{WSOD2}^{+}$ \cite{Zeng_2019_ICCV} & 87.1 & 80.0 & 74.8 & 60.1 & 36.6 & 79.2 & 83.8 & 70.6 & 43.5 & {88.4} & 46.0 & {74.7} & 87.4 & 90.8 & 44.2 & 52.4 & 81.4 & 61.8 & 67.7 & 79.9 & 69.5\\
    Arun~ \textit{et al.}\cite{Arun_2019} & {88.6} & {86.3} & 71.8 & 53.4 & 51.2 & {87.6} & {89.0} & 65.3 & 33.2 & 86.6 & 58.8 & 65.9 & {87.7} & {93.3} & 30.9 & 58.9 & 83.4 & 67.8 & 78.7 & 80.2 & {70.9} \\
    $\text{MIST (Full)}^{+}$ \cite{ren-wetectron2020} & 87.5 & 82.4 & {76.0} & 58.0 & 44.7 & 82.2 & 87.5 & 71.2 & {49.1} & 81.5 & 51.7 & 53.3 & 71.4 & 92.8 & 38.2 & 52.8 &79.4 & 61.0 & 78.3 & 76.0  & 68.8 \\
    \hline
    WSLAT-Ens \cite{wslat2015} & 78.6 & 63.4 & 66.4 & 56.4 & 19.7 & 82.3 & 74.8 & 69.1 & 22.5 & 72.3 & 31.0 & 63.0 & 74.9 & 78.4 & 48.6 & 29.4 & 64.6 & 36.2 & 75.9 & 69.5  & 58.8 \\
    $\text{MSD-Ens}^{+}$ \cite{msd2018} & 89.2 & 75.7 & 75.1 & 66.5 & 58.8 & 78.2 & 88.9 & 66.9 & 28.2 & 86.3 & 29.7 & 83.5 & 83.3 & 92.8 & 23.7 & 40.3 & 85.6 & 48.9 & 70.3 & 68.1 & 66.8 \\
    OICR+UBBR \cite{ubbr2018} & 47.9 & 18.9 & 63.1 & 39.7 & 10.2 & 62.3 & 69.3 & 61.0 & 27.0 & 79.0 & 24.5 & 67.9 & 79.1 & 49.7 & 28.6 & 12.8 & 79.4 & 40.6 & 61.6 & 28.4 & 47.6 \\
    \hline
    \textbf{Ours} & 89.6 & 82.0 & 73.6 & 55.3 & 48.9 & 86.3 & 87.3 & 83.1 & 45.3 & 87.7 & 48.3 & 82.3 & 80.6 & 90.8 & 36.3 & 52.0 & 88.7 & 66.1 & 81.7 & 80.3 & \textbf{72.3} \\
    $\textbf{Ours}^{+}$ & 89.2 & 82.0 & 74.2 & 53.2 & 51.2 & 84.8 & 87.5 & 83.7 & 46.2 & 87.0 & 48.3 & 84.7 & 79.9 & 92.4 & 40.3 & 47.6 & 88.7 & 65.6 & 81.0 & 81.7 & \textbf{72.5}   \\
    \specialrule{.15em}{.05em}{.05em}
    \textbf{Upper bounds:} \\
    \hline
    Faster R-CNN \cite{renNIPS15fasterrcnn} & 99.6 & 96.1 & 99.1 & 95.7 & 91.6 & 94.9 & 94.7 & 98.3 & 78.7 & 98.6 & 85.6 & 98.4 & 98.3 & 98.8 & 96.6 & 90.1 & 99.0 & 80.1 & 99.6 & 93.2 & 94.3 \\
    $\text{Zhong \textit{et al.} (R50-C4)}^{*}$ \cite{zhong2020boosting} & 86.7 & 62.4 & 87.1 & 70.2 & 66.4 & 85.3 & 87.6 & 88.1 & 42.3 & 94.5 & 32.3 & 87.7 & 91.2 & 88.8 & 71.2 & 20.5 & 93.8 & 51.6 & 87.5 & 76.7 & 73.6 \\
    $\text{Zhong \textit{et al.} (R50-C4)}^{+*}$ \cite{zhong2020boosting} & 87.5 & 64.7 & 87.4 & 69.7 & 67.9 & 86.3 & 88.8 & 88.1 & 44.4 & 93.8 & 31.9 & 89.1 & 92.9 & 86.3 & 71.5 & 22.7 & 94.8 & 56.5 & 88.2 & 76.3 & 74.4 \\
    \specialrule{.15em}{.05em}{.05em}
    \end{tabular}
    }}
    \label{table:per-cls-voc07-corloc}
    \end{table*}
    
    \begin{table*}[tbh]
        \caption{Single model correct localization (CorLoc) results on PASCAL VOC  2012, where $\text{~}^{+}$ means the results with multi-scale testing, $\text{~}^{*}$ means joint training of WSOD models on the auxiliary dataset and weakly-annotated dataset.}
    \centering
    \resizebox{\textwidth}{!}{
    {
    \begin{tabular}{l | c c c c c c c c c c c c c c c c c c c c | c}
    \specialrule{.15em}{.05em}{.05em}
    Methods & Aero & Bike & Bird & Boat & Bottle & Bus & Car & Cat & Chair & Cow & Table & Dog & Horse & Motor & Person & Plant & Sheep & Sofa & Train & ~~TV~~ & CorLoc \\
    \hline
    $\text{OICR}^{+}$ \cite{tang2017multiple} &-&-&-&-&-&-&-&-&-&-&-&-&-&-&-&-&-&-&-&-& 62.1 \\
    $\text{PCL}^{+}$ \cite{tang2018pcl} & 77.2 & 83.0 & 62.1 & 55.0 & 49.3 & 83.0 & 75.8 & 37.7 & 43.2 & 81.6 & 46.8 & 42.9 & 73.3 & 90.3 & 21.4 & 56.7 & 84.4 & 55.0 & 62.9 & 82.5 & 63.2 \\
    Shen~\cite{Shen_2019_CVPR} &-&-&-&-&-&-&-&-&-&-&-&-&-&-&-&-&-&-&-&-& 63.5 \\
    $\text{Li}^{+}$ \cite{li2019weakly} &-&-&-&-&-&-&-&-&-&-&-&-&-&-&-&-&-&-&-&-& 67.9 \\
    $\text{C-MIL}^{+}$ \cite{Wan_2019} &-&-&-&-&-&-&-&-&-&-&-&-&-&-&-&-&-&-&-&-& 67.4 \\
    $\text{Yang }\textit{et al.}^{+}$ \cite{yang2019towards} & 82.4 & 83.7 & {72.4} & {57.9} & 52.9 & 86.5 & 78.2 & {78.6} & 40.1 & 86.4 & 37.9 & {67.9} & {87.6} & 90.5 & 25.6 & 53.9 & 85.0 & {71.9} & 66.2& 84.7 & 69.5 \\
    Arun~ \textit{et al.}\cite{Arun_2019} &-&-&-&-&-&-&-&-&-&-&-&-&-&-&-&-&-&-&-&-& 69.5 \\
    $\text{WSOD2}^{+}$ \cite{Zeng_2019_ICCV} &-&-&-&-&-&-&-&-&-&-&-&-&-&-&-&-&-&-&-&-& {71.9} \\
    $\text{MIST (Full)}^{+}$ \cite{ren-wetectron2020} & {91.7} & {85.6} & 71.7 & 56.6 & {55.6} & {88.6} & 77.3 & 63.4 & {53.6} & {90.0} & 51.6 & 62.6 & 79.3 & {94.2} & 32.7 & 58.8 & {90.5} & 57.7 & {70.9} & {85.7} & 70.9 \\
    \hline
    \textbf{Ours} & 91.9 & 87.4 & 81.9 & 66.7 & 58.5 & 91.2 & 79.9 & 67.3 & 50.0 & 91.9 & 49.6 & 80.3 & 89.6 & 91.8 & 15.6 & 58.8 & 88.7 & 67.1 & 70.2 & 85.0 & \textbf{73.2} \\ 
    $\textbf{Ours}^{+}$ & 91.9 & 87.2 & 81.0 & 66.9 & 61.3 & 90.7 & 81.2 & 66.8 & 51.2 & 91.9 & 50.4 & 81.0 & 90.5 & 91.4 & 16.1 & 58.5 & 89.9 & 67.8 & 70.0 & 86.7 & \textbf{73.7} \\
    \specialrule{.15em}{.05em}{.05em}
    \end{tabular}
    }}
    \label{table:per-cls-voc12-corloc}
    \end{table*}

\begin{table*}[tbh]
    \caption{Detailed comparison of different methods on COCO-20.}
        \centering
            \footnotesize{
        \begin{tabular}{l |  c c c c c c c c c c}
        \specialrule{.15em}{.05em}{.05em}
        Methods  &  mAP  & AP50 & AP75 & $AP_S$ & $AP_M$ & $AP_L$ & $AR_{100}$ & $AR_S$ & $AR_M$ & $AR_L$ \\
        \hline
        OICR  & 9.5 & 22.8 & 6.8 & 2.4 & 9.4 & 17.5 & 24.2 & 8.0 & 21.8 & 38.9 \\
        OICR+REG  & 10.4 & 23.9 & 8.1 & 3.9 & 9.5 & 17.8 & 22.3 & 7.5 & 19.3 & 35.1 \\
        \hline
        Ours LBBA  & {13.0} & {27.5} & 11.2 & 4.1 & 12.5 & 21.4 & 25.1 & 8.6 & 23.3 & 38.4 \\
        Ours LBBA+masking  & {13.7} & {29.9} & 11.5 & 4.2 & 13.0 & 22.1 & 25.8 & 8.8 & 23.9 & 39.7 \\
        \specialrule{.15em}{.05em}{.05em}
        \end{tabular}
        }
        \label{table:coco20}
    \end{table*}

\begin{table}[t]
   \caption{Experimental settings on auxiliary datasets and target datasets. }
   \centering
      \footnotesize{
         \begin{tabular}{l |  c c }
            \specialrule{.15em}{.05em}{.05em}
            Data Settings & Auxiliary Datasets & Target Datasets\\
            \hline
            Setting 1 & COCO-60 & PASCAL VOC 2007  \\
            Setting 2 & COCO-60 & PASCAL VOC 2012  \\
            Setting 3  & COCO-60 & COCO-20\\
            Setting 4 & ILSVRC-Source& ILSVRC-Target \\
            \specialrule{.15em}{.05em}{.05em}
         \end{tabular}
      }
      \vspace{-1em}
      \label{table:a-pair-data}
   \end{table} 

\section{Datasets}
\label{sec:datasets}
To illustrate the effectiveness of our method, we conduct experiments on various representative datasets: PASCAL VOC 2007 and 2012 datasets, MS-COCO dataset, and ILSVRC 2013 detection dataset. 
\subsection{Auxiliary Datasets}
\paragraph{COCO-60 Dataset}
MS-COCO 2017~\cite{lin2014microsoft} is a large-scale object detection dataset. Note that MS-COCO dataset includes 80 different object classes. To eliminate semantic overlap and show the generalization ability of our method, we construct a subset of MS-COCO by excluding PASCAL VOC classes instance annotations and call it COCO-60. As such, COCO-60 dataset contains $\sim$98K training images and $\sim$4K validation images, respectively. Construction details are shown as Appendix~\ref{sec:coco6020}.
\paragraph{ILSVRC-Source Dataset}
To prove that our method can be generalized to more categories, we conduct extended experiments on the ILSVRC2013 detection dataset.
ILSVRC detection dataset contains 200 categories, which is much more than that for PASCAL VOC or COCO-20. To construct the corresponding auxiliary dataset, we select the first 100 classes sorted in alphabetic order as the source classes in the auxiliary dataset. Construction details are shown as Appendix~\ref{sec:imagenetST}.
\subsection{Target Datasets}
\paragraph{PASCAL VOC Dataset}
PASCAL VOC 2007 and 2012 datasets contain 9,963 images and 22,531 images collected from 20 object classes, respectively. For fair comparison, we use \textit{trainval} set for training WSOD networks and report evaluation results on \textit{test} set. During the training process, only image-level labels are used as supervision.
\paragraph{COCO-20 Dataset}
To verify the generalization ability of our LBBA, we construct another target dataset from MS-COCO dataset namely COCO-20 dataset. Note that the COCO-20 dataset has the same 20 classes as PASCAL VOC dataset, but containing more complicated scenarios in images. Construction details are shown as Appendix~\ref{sec:imagenetST}.
\paragraph{ILSVRC-Target Dataset}
ILSVRC detection dataset contains 200 categories. To construct the target dataset and avoid semantic overlaps with the corresponding auxiliary dataset, we select the last 100 classes sorted in alphabetic order as target classes in our weakly supervised object detection dataset. Construction details are shown as Appendix~\ref{sec:imagenetST}.
\subsection{Auxiliary-Target Pairs}
From these datasets, we divide them into four dataset-pair settings, an auxiliary dataset corresponding to a target dataset, to deploy experiments.
Table~\ref{table:a-pair-data} give the dataset-pair settings.
Setting 1 and Setting 2 are mentioned in section 4 of main paper and we will introduce details of setting 3 and setting 4 in Appendix~\ref{sec:coco6020} and Appendix~\ref{sec:imagenetST}. Then we will state more experimental results in Appendix~\ref{sec:coco20} and Appendix~\ref{sec:imagenetT}. 

\subsection{Construction of COCO-60/COCO-20}
\label{sec:coco6020}
To simplify the statement, we define COCO-60 classes as the categories in original COCO classes but excluding PASCAL VOC classes. Then we state how to construct COCO-60 dataset and COCO-20 dataset. 

To construct COCO-60 dataset, we first keep annotations of COCO-60 classes in COCO 2017 \textit{train} set, then we select images which contain at least one instance of COCO-60 classes in COCO 2017 \textit{train} set to construct our COCO-60 \textit{train} set. Next we keep the same steps to build up our COCO-60 \textit{val} set.

Besides, we also follow Zhong \etal~\cite{zhong2020boosting} to define a COCO-60-clean dataset. Particularly, we select images which \textbf{only contain instances of COCO-60 classes} in COCO 2017 \textit{train} set to construct COCO-60-clean \textit{train} set, and obtain only 21987 training images.
Compared to COCO-60 dataset, COCO-60-clean dataset does not exist objects of VOC classes in the background of images, such that this dataset is cleaner than our COCO-60 dataset and easier to learn. 
We will discuss the difference between our method and Zhong \etal~\cite{zhong2020boosting} based on COCO-60 and COCO-60-clean datasets. 

As for COCO-20 dataset, we select images which only contain instances of 20 PASCAL VOC classes in COCO 2017 \textit{train} set to construct our COCO-20 \textit{train} set. Next we keep annotations of 20 PASCAL VOC classes in COCO 2017 \textit{val} set, and then select images which contain at least one instance of 20 PASCAL VOC classes in COCO 2017 \textit{val} set to construct our COCO-20 \textit{val} set.

\subsection{Construction of ILSVRC-Source/Target}
\label{sec:imagenetST}
The original ILSVRC dataset contains a training set and a validation set. 
Firstly, We split the validation set into val1 validation set and val2 validation set.
Then we state how to construct ILSVRC-Source dataset and ILSVRC-Target dataset.

To construct ILSVRC-Source training set,  we keep images of the first 100 categories sorted in alphabetic order from val1 and sample 1000 images per category in the same 100 categories from ILSVRC training set as data augmentation.

To construct ILSVRC-Target training set,  we keep images of the latter 100 categories sorted in alphabetic order from val1 and sample a maximum of 1000 images per category in latter categories from ILSVRC training set to augment it,while keeping only image-level labels. And to construct ILSVRC-Target test set, we keep images of the same 100 categories from val2.
\section{Implementation Details}
\label{sec:impl}
\subsection{Overall Implementation Details}
For LBBAs, we apply Faster R-CNN~\cite{renNIPS15fasterrcnn} with backbone of ResNet-50~\cite{He2016DeepRL} and we adopt class-agnostic bounding box adjusters to eliminate potential semantic information leak in bounding box refinement. For WSOD network, we apply OICR~\cite{tang2017multiple} with a backbone of VGG-16~\cite{simonyan2014very} and introduce a class-agnostic bounding box regression branch. 
Following the settings of \cite{bilen2016weakly,tang2017multiple,tang2018pcl,yang2019towards,ren-wetectron2020,Arun_2019,zhong2020boosting}, we initialize backbone models of two networks with ImageNet~\cite{deng2009imagenet} pre-trained weights while other layers are randomly initialized. 
As suggested in \cite{ren-wetectron2020,yang2019towards,tang2018pcl,tang2017multiple,Zeng_2019_ICCV}, we use MCG boxes as precomputed proposals for COCO-60 and use Selective Search boxes as precomputed proposals for PASCAL VOC.
During training, both two networks are optimized by stochastic gradient descent (SGD) with the batch size of 1 and initialized learning rate of 0.001.
In each stage, LBBA is trained with 4 epochs, and the learning rate is decayed by 0.1 after 3 epochs. 
Analogously, WSOD network is trained within 20 epochs and learning rate is decayed by 0.1 after 10 epochs. 
All programs are implemented by PyTorch toolkit, and all experiments are conducted on a single NVIDIA RTX 2080Ti GPU.

For the multi-label image classifier, we adopt the ADD-GCN \cite{ye2020add}, which builds a Dynamic Graph Convolutional Network (D-GCN) to model the relation of content-aware category representations generated by a Semantic Attention Module(SAM). 
During training, the ADD-GCN is optimized by SGD with batch size of 16. 
The learning rate is initially set to 0.05 for training 40 epoch and decayed by 0.1 to train the latter 10 epoch. 
The best threshold $\tau$ is set to -3.0. 
By the way, the setting of the $\tau$ is based on the implementation of multi-label image classifier.
Too high or too low will be detrimental to the final result, and we will give the results and analysis in the next section.

All the source code and pre-trained models will be made publicly available.

\subsection{Structure of LBBA}
Here we briefly introduce the structure of LBBA. 
In our solution, we adopt Faster R-CNN~\cite{renNIPS15fasterrcnn} with backbone of ResNet-50~\cite{He2016DeepRL} as our LBBA. And LBBA is designed to be a class-agnostic bounding box regressor to eliminate potential semantic information leak in bounding box refinement. 
Note that the inside RPN~\cite{renNIPS15fasterrcnn} is only used during EM-like LBBA training to improve the training stabilization and generalization ability of LBBA, and will not be used during the inference stage.
We argue that using Faster R-CNN as adjuster has two merits. 
(i) For the initialization of LBBA training, Faster R-CNN exhibits better performance than Fast R-CNN. 
(ii) By combining precomputed proposals and proposals from RPN, box regression branch of LBBA can generalize better to various proposals, resulting in more precise box refinement results.

\begin{table*}[tbh]
    \caption{Per-class mIoU and average mIoU of our LBBA with precomputed proposals. It is clear to conclude that LBBA obtains more precise box refinement ability.}
\centering
\resizebox{\textwidth}{!}{
{
\begin{tabular}{l | c c c c c c c c c c c c c c c c c c c c | c}
\specialrule{.15em}{.05em}{.05em}
Method & Aero & Bike & Bird & Boat & Bottle & Bus & Car & Cat & Chair & Cow & Table & Dog & Horse & Motor & Person & Plant & Sheep & Sofa & Train & ~~TV~~ & mIoU \\
\hline
Precomputed Proposals &46.1&45.7&45.3&45.3&44.6&46.1&45.7&47.1&45.8&45.6&48.6&46.2&45.8&46.1&45.5&45.0&45.0&47.8&46.9&45.0& 45.9 \\
LBBA Module & 63.0 & 54.6 & 65.5 & 60.8 & 60.5 & 68.3 & 68.3 & 69.4 & 57.4 & 69.8 & 57.8 & 69.0 & 65.6 & 58.7 & 59.3 & 52.0 & 66.7 & 64.5 & 66.0 & 68.5 & 63.2 \\
\specialrule{.15em}{.05em}{.05em}
\end{tabular}
}}
\label{table:miou}
\end{table*}

\begin{table}[t]
   \caption{Does ideal LBBA improve performance of WSOD?}
       \centering
          \footnotesize{
       \begin{tabular}{l |  c }
       \specialrule{.15em}{.05em}{.05em}
       Methods & mAP (VOC07) \\
       \hline
       baseline OICR+\cite{ren-wetectron2020} & 51.4 \\
       Ours LBBA  & 55.8  \\
       Ours LBBA (ideal) & 58.4  \\
       \specialrule{.15em}{.05em}{.05em}
       \end{tabular}
       }
       \label{table:ideallbba}
       \vspace{-1em}
   \end{table}
\section{More Ablation Studies}
\label{sec:ablation_more}
\subsection{Evaluating LBBA Module Separately}
In our solution, LBBA module is designed to be class-agnostic, making that the learned box regressors can be shared among different object classes and transfered to newly added classes. Though we have shown the positive effect of LBBA module in terms of mAP metric, we still evaluate it separately in a manner of proposal evaluation. Therefore we calculate mean IoU between refined proposals from LBBA module and GT boxes. As a comparison, we also calculate mIoU between precomputed proposals and GT boxes as a baseline. IoU performance of LBBA is shown as Table \ref{table:miou}. It is clear to conclude that our LBBA module obtains more precise box refinement ability after EM-like LBBA training.
\subsection{Performance with ideal LBBA}
Our observation is that localization attribute is shared among all kinds of objects, such that a fully supervised box refinement network trained on an auxiliary dataset can be utilized during transfer learning. 
Therefore, to verify our observation, we build another LBBA-boosted WSOD experiment. 
During this experiment, we replace pretrained LBBA network by ground-truth bounding box and keep using image class labels to supervise MIL branch, 
because ground-truth boxes can be seen as an ideal LBBA network to supervise box regression branch of WSOD network during LBBA-boosted WSOD.
And then we execute such LBBA-boosted WSOD with the same training schedule. 
Detection performance of WSOD with ideal LBBA on PASCAL VOC 2007 \textit{test} set is shown as Table~\ref{table:ideallbba}.
Compared to baseline OICR~+\cite{ren-wetectron2020} as well as our proposed LBBA, LBBA-boosted WSOD with ideal LBBA ourperforms by 7.0\% on mAP and 2.6\% on mAP, respectively. 
This improvement verifies our observation, and also encourages us to develop more effective adjusters. 
\subsection{Effect of Masking Strategy for Proposal Classification}
Improving the performance of proposal classification usually benefits to improving the overall detection performance of WSOD. Therefore, we also explore the effect of our masking strategy in our LBBA-boosted WSOD network.
To demonstrate the effect of the masking strategy, we compared LBBA method with masking strategy with pure LBBA.
Table \ref{table:add-gcn} shows the effect of the masking strategy of proposal classification. 
Compared to pure LBBA with OICR and OICR~+\cite{ren-wetectron2020}, our masking strategy improves detection performance by 1.3\% and 0.7\% mAP on PASCAL VOC 2007 \textit{test} set.
We also explore the effect of $\tau$ in masking strategy, experimental result is shown as Table~\ref{table:diff-tau}, we found that $\tau=-3.0$ is the best selection during our masking strategy. Above results indicate that classification predictions from multi-label image classifier are able to select categories with high scores.
By suppressing the bounding box scores of non-appearing categories, the proportion of false positives in the final test results is reduced, which is beneficial to improving the overall detection performance of WSOD.

\begin{table}[t]
\caption{Does one-class LBBA improve performance of WSOD?}
    \centering
       \footnotesize{
    \begin{tabular}{l |  c }
    \specialrule{.15em}{.05em}{.05em}
    Methods & mAP (VOC07) \\
    \hline
    Ours LBBA  & 55.8  \\
    Ours LBBA (one class) & 56.2  \\
    \specialrule{.15em}{.05em}{.05em}
    \end{tabular}
    }
    \label{table:oclbba}
    \vspace{-1em}
\end{table}

\begin{table}[t]
   \caption{Detailed comparison of different methods on ILSVRC13 Target}
       \centering
          \footnotesize{
       \begin{tabular}{l |  c }
       \specialrule{.15em}{.05em}{.05em}
       Methods & AP50 \\
       \hline
       OICR  & 20.5  \\
       OICR+REG & 22.4  \\
       \hline
       LBBA(OICR) & 28.0  \\
       LBBA(OICR)+masking & 30.1 \\
       \specialrule{.15em}{.05em}{.05em}
       \end{tabular}
       }
       \label{table:imagenetlbba}
       \vspace{-1em}
   \end{table}

\subsection{Is One-class Adjuster Necessary?}
During our experiments, to simplify overall experimental settings, 
we adopt conventional Faster R-CNN~\cite{renNIPS15fasterrcnn} with class-agnostic box regression branch as our LBBA fundamental structure, 
and keep the original RoI classification branch (\textit{e.g.}, 60 classes on COCO-60 dataset). 
But how the performance of LBBA-boosted WSOD will be changed if we use class-agnostic detector as our LBBA? 
To solve this question, we train another LBBA whose box regression branch and RoI classification branch are both class-agnostic. 
And then we execute EM-like LBBA training as well as LBBA-boosted WSOD sequentially using one-class LBBA mentioned above.
Performance of LBBA-boosted WSOD supervised by one-class LBBA on PASCAL VOC 2007 is shown as Table~\ref{table:oclbba}. 
Compared to WSOD with our proposed standard LBBA, LBBA with one-class LBBA achieves a slight performance improvement (56.2\% mAP vs. 55.8\% mAP) on PASCAL VOC 2007 \textit{test} set. 
However, using conventional LBBA during our experiment is convenient and flexible because each pretrained object detection network can be utilized as a pretrained LBBA directly. 
Based on this observation, we keep using conventional Faster R-CNN~\cite{renNIPS15fasterrcnn} as our LBBA.

\begin{table}[t]
   \caption{Comparison of updating pipeline of $f$ with $\theta_{f}$ (here we set T=3). Evaluation result shows that updating progressively achieves better performance while updating with last $g_{T}$ achieves a similar performance with only one training stage.}
       \centering
          \footnotesize{
       \begin{tabular}{l | c | c }
       \specialrule{.15em}{.05em}{.05em}
       Methods & Stages & mAP (VOC07) \\
       \hline
       updating progressively & 4 & 55.8  \\
       updating with last $g_{T}$ & 1 & 55.4  \\
       \specialrule{.15em}{.05em}{.05em}
       \end{tabular}
       }
       \label{table:training_procedure}
       \vspace{-1em}
   \end{table}

\subsection{How to update $\theta_{f}$?}
During our LBBA-boosted WSOD in Sec.~{3}, we use $\{{g}_{{0}}\dots{g}_{{T}}\}$ with corresponding parameters $\{{\theta}_{g}^{{0}}\dots{\theta}_{g}^{{T}}\}$ to supervise our WSOD network $f$ with $\theta_{f}$ progressively. And to construct a simpler training pipeline, we can directly use the last $g_{T}$ to supervise $f$ with $\theta_{f}$. Therefore we are curious about the performance gap between updating $\theta_{f}$ progressively and updating $\theta_{f}$ directly. Corresponding evaluation results are shown as Table~\ref{table:training_procedure}. The WSOD network updated progressively achieves better performance, while the WSOD network updated with the last $g_{T}$ achieves a similar performance (-0.4\% in terms of mAP on VOC 2007 dataset) with only one training stage. This result indicates that we can build a lighter LBBA-boosted WSOD training pipeline by only using the last $g_{T}$ in practice, but training progressively is usually stable and better. 

\section{Comparison with State-of-the-arts}
\begin{figure*}
    \begin{center}
    \includegraphics[width=6in]{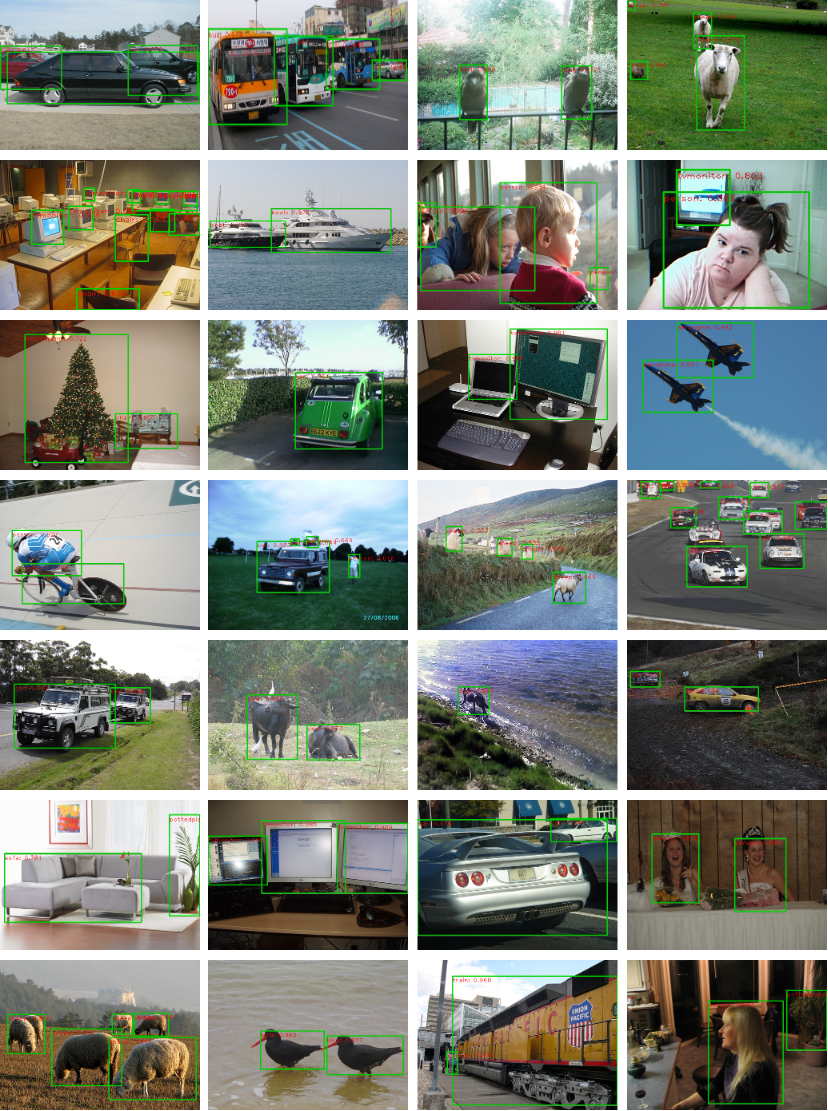}
    \end{center}
        \caption{More visualization results of our method on PASCAL VOC 2007, which has the ability to generate precise bounding boxes.}
    \label{fig:vis-2007-onepage}
    \end{figure*}

\begin{figure*}
    \begin{center}
    \includegraphics[width=6in]{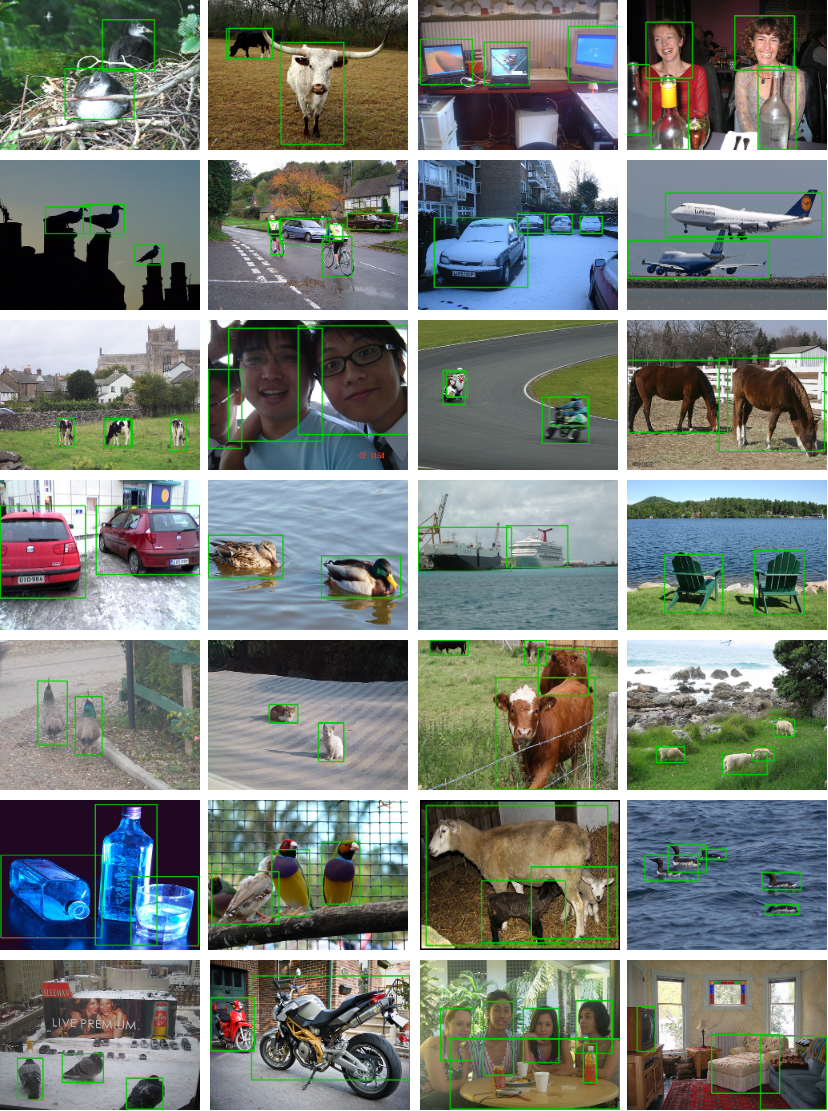}
    \end{center}
        \caption{More visualization results of our method on PASCAL VOC 2012, which has the ability to generate precise bounding boxes.}
    \label{fig:vis-2012-onepage}
    \end{figure*}

We compare our method with several state-of-the-art WSOD approaches in terms of detection and localization performance on PASCAL VOC datasets. As suggested in \cite{bilen2016weakly,tang2017multiple,tang2018pcl,yang2019towards,ren-wetectron2020,Arun_2019,zhong2020boosting}, we report detection results on \textit{test} set and localization results on \textit{trainval} set, respectively. 
Table~\ref{table:per-cls-voc07} and Table~\ref{table:per-cls-voc12} compares the results of different state-of-the-art WSOD approaches on PASCAL VOC 2007 and 2012 datasets. It can be seen that our LBBA improves OICR and OICR+REG over 15.3\% and 5.0\% on PASCAL VOC 2007 dataset, respectively. Furthermore,  our method performs better than all competing methods, except Zhong \textit{et al.}~\cite{zhong2020boosting}. Note that \cite{zhong2020boosting} uses a stronger backbone model and knowledge transfer strategy by directly incorporating source and target datasets. As shown in Fig.~\ref{fig:vis-2007-onepage}, our method has the ability to generate precise bounding boxes. On PASCAL VOC 2012, our LBBA is superior to all competing methods and obtains more than 1\% gains over all WSOD approaches. Experimental results show that our method is effective in improving the detection performance of WSOD. As shown in Fig.~\ref{fig:vis-2012-onepage}, our method also has the ability to generate precise bounding boxes on PASCAL VOC 2012 dataset.

We further evaluate the localization performance of our method. Table~\ref{table:per-cls-voc07-corloc} and Table~\ref{table:per-cls-voc12-corloc} lists the results of several state-of-the-art WSOD approaches on PASCAL VOC 2007 and 2012. Our LBBA outperforms OICR by 11.7\% and also improves the baseline OICR+REG over 4.3\% on PASCAL VOC 2007 dataset. Besides, our LBBA performs better than all competing methods. Meanwhile, on PASCAL VOC 2012, our LBBA is also superior to all competing methods and obtains 1.3\% over WSOD 2\cite{Zeng_2019_ICCV}.
In comparison to Zhong \etal \cite{zhong2020boosting}, our LBBA-based method employs a weaker backbone model and avoids the direct joint use of the source and target datasets, while still achieving competitive CorLoc results under the settings of both single-scale testing and multi-scale testing. 
Above results show that our LBBA-based method is effective in improving the localization performance of WSOD.

\section{Generalization to COCO-20}
\label{sec:coco20}
We verify the generalization ability of our LBBA method using a COCO-20 dataset. To this end, we build COCO-20 dataset by collecting the images that only contain instances belonging to the remain 20 classes from \textit{train} and \textit{val} sets of COCO 2017 \cite{lin2014microsoft}, and use them as the corresponding \textit{train} and \textit{val} sets. Comparing with PASCAL VOC, COCO-20 is more challenging due to more instances and complex layouts. Here we adopt OICR+REG as WSOD network $f$, and compare with OICR and OICR+REG as baseline methods. We train all models using exactly the same settings in sec.~\ref{sec:impl}, and the results are listed in Table \ref{table:coco20}. 
Note that our LBBA method with masking strategy outperforms OICR and OICR+REG by 3.5\% (4.7\%) and 2.6\% (3.6\%) in terms of mAP and AP50, clearly demonstrating the generalization ability of our LBBA method.
After adding masking strategy, our LBBA method outperforms OICR and OICR+REG by 4.2\% (7.1\%) and 3.3\% (6.0\%) in terms of mAP and AP50, which demonstrates the effectiveness of our masking strategy.

\section{Generalization to ILSVRC-Target}
\label{sec:imagenetT}
To illustrate that our method can be generalized to more categories, we build the ILSVRC-Target dataset following Appendix~\ref{sec:imagenetST} and conduct experiments on it.
The baseline models setting is same as Appendix~\ref{sec:coco20} and results are listed in Table~\ref{table:imagenetlbba}.
Note that our LBBA method outperforms OICR and OICR+REG by 7.5\% and 5.6\% in terms of AP50, which proves that our method can withstand the test of scenes containing more categories of objects.
Furthermore, with the enhancement of masking strategy, the performance of WSOD network further outperforms pure LBBA-boosted WSOD by 2.1\% in terms of AP50, which shows that masking strategy is able to improve quality of proposal classification and can be generalized to more categories simultaneously.
\begin{table}
   \caption{Effect of Masking Strategy, where \emph{+masking} means our LBBA with masking strategy.}
      \centering
         \footnotesize{
      \begin{tabular}{l |  c }
      \specialrule{.15em}{.05em}{.05em}
      Methods & mAP (VOC07) \\
      \hline
      LBBA(OICR) & 55.1  \\
      LBBA(OICR)+masking & 56.4  \\
      \hline
      LBBA(OICR+\cite{ren-wetectron2020}) & 55.8  \\
      LBBA(OICR+\cite{ren-wetectron2020})+masking & 56.5  \\
      \specialrule{.15em}{.05em}{.05em}
      \end{tabular}
      }
      \label{table:add-gcn}
      \vspace{-1em}
\end{table}

\begin{table}
   \caption{Varying $\tau$ for Multi-Label Image Classifier. We evaluated $\tau$ on LBBA-Boosted WSOD with OICR head.}
      \centering
         \footnotesize{
      \begin{tabular}{l |  c }
      \specialrule{.15em}{.05em}{.05em}
      $\tau$ & mAP (VOC07) \\
      \hline
      +0.5 & 55.4  \\
      -0.5 & 55.7  \\
      -1.5 & 56.1  \\
      $\textbf{-3.0}$ & $\textbf{56.4}$  \\
      -6.0 & 56.3  \\
      -10.0 & 56.1  \\
      -12.0 & 55.8  \\
      -20.0 & 55.3  \\
      \specialrule{.15em}{.05em}{.05em}
      \end{tabular}
      }
      \label{table:diff-tau}
      \vspace{-1em}
\end{table}

\section{Discussion}
In this section, we will discuss our proposed LBBA as well as some modern weakly supervised object detection algorithms in different aspects.
\subsection{Discussion of our LBBA}
\label{sec:discuss_lbba}
Here we discuss several potential merits of the problem setting and our proposed method.
In LBBA-boosted WSOD, the auxiliary well-annotated dataset is not needed and only a smaller amount (\eg, 3) of LBBAs are required.
Thus, our problem setting allows deploying LBBAs to versatile weakly annotated datasets for boosting detection performance while avoiding the leakage of well-annotated dataset.
In terms of memory consumption, LBBAs are much more economical than the storage of well-annotated dataset.

For the sake of generalization ability, we adopt class-agnostic LBBAs.
In comparison to the universal bounding box regressor~\cite{ubbr2018}, stage-wise LBBAs are specifically learned to adjust the region proposals generated by WSOD towards the ground-truth bounding boxes, and thus are more effective.
To show the generalization ability, the LBBAs learned from well-annotated dataset can be readily deployed to the weakly-annotated dataset with non-overlapped object classes.
Nonetheless, LBBAs also work well when the weakly-annotated dataset has the overlapped object classes.

Furthermore, the two subtasks, \ie, learning bounding box adjusters and LBBA-boosted WSOD, can be respectively regarded as a kind of knowledge extraction and transfer.
With learning bounding box adjusters, we extract the knowledge from the auxiliary well-annotated dataset.
Consequently, the extracted knowledge, \ie, LBBAs, will be transferred to the WSOD models for improving detection performance.
In comparison to directly incorporating auxiliary dataset with weakly-annotated dataset, we argue that the separation of knowledge extraction and transfer is practically more natural, convenient, and acceptable.
\subsection{Discussion of \emph{ResNet-WS}}
Shen \textit{et al.} \cite{DRN-WSOD_2020_ECCV} proposed a novel residual network backbone architecture, which combines the advantage of residual blocks for feature extraction as well as redundant adaptation neck like \emph{fc6-fc7} of VGG, and leads to better detection performance of the residual network with the weakly supervised setting. 

Due to hardware limitations, we did not employ ResNet-WS backbone in our experiments. However, such improvements mainly focus on the backbone of WSOD networks and are able to easily plug into our framework to improve the overall performance of our proposed method. We believe that such method is compatible with ours.
\subsection{Discussion of \emph{CASD}}
Recently we noticed that Huang \textit{et al.} \cite{huangCASD2020} proposed a novel \textit{Comprehensive Attention Self-Distillation} approach to further improve performance of weakly supervised object detection. This approach obtains higher detection performance than ours and lower localization performance than ours. Similarly, as mentioned in the ablation study, our approach is compatible with various WSOD heads. Naturally, CASD is also compatible. We also believe that the detection performance of WSOD can be better when we apply CASD to our proposed method.

\subsection{Discussion of Zhong \etal}
\begin{table}[t]
   \caption{Some analysis of Zhong \etal in iteration 0. We keep auxiliary dataset and weakly annotated dataset isolated to evaluate performance of Zhong \etal fairly.}
       \centering
          \footnotesize{
       \begin{tabular}{l |  c }
       \specialrule{.15em}{.05em}{.05em}
       Methods & mAP (VOC07) \\
       \hline
       Zhong \etal~\cite{zhong2020boosting} iter 0  & 54.4  \\
       Zhong \etal~\cite{zhong2020boosting} w/o Test-Time Aug iter 0 & 41.8  \\
       Zhong \etal~\cite{zhong2020boosting} w/ COCO-60-full iter 0  & $\sim$45  \\
       \specialrule{.15em}{.05em}{.05em}
       \end{tabular}
       }
       \label{table:wsod_transfer}
       \vspace{-1em}
   \end{table}
Zhong \etal proposed a novel transfer learning based weakly supervised object detection framework, which utilizes a progressive knowledge distillation training procedure and builds up a universal object proposal generator as well as the corresponding WSOD network. 

This method achieves the state-of-the-art detection performance on PASCAL VOC dataset. However, ththisese method exists some difference with our proposed method, which can be listed as follows. First, the Method of Zhong \etal proposed a kind of proposal generator while our proposed method is a kind of box refinement network. Second, during EM-like Multi-stage LBBA training as well as LBBA-boosted WSOD, we keep auxiliary dataset and weakly annotated dataset isolated to avoid information leakage of weakly annotated dataset. Finally, after LBBA-boosted WSOD, our WSOD network can generate object detection results individually without help from LBBA. 

Besides, the approach of Zhong \etal also suffers from \emph{three fundamental limitations during applications}. \textbf{First}, when training OCUD in iteration 1 or 2, ground-truth data from auxiliary dataset and pseudo labels from weakly annotated detection dataset are mixed and fed into the OCUD network jointly. As we discussed in Section~\ref{sec:discuss_lbba}, this mixture might introduce information leakage of weakly annotated dataset and longer training time in practice. 

\textbf{Second}, to improve detection performance during evaluation, predictions from the MIL network of Zhong \etal are augmented by adding corresponding objectness scores from OCUD. When removing \emph{Test-Time-Augmentation} (same with using MIL network individually), the performance of Zhong \etal drops to 41.8\% mAP. 

\textbf{Finally}, Zhong \etal~\cite{zhong2020boosting} trains the OCUD on COCO-60-clean dataset which is mentioned in Sec.~\ref{sec:coco6020}, and this dataset is easier to learn. Different from \cite{zhong2020boosting}, we optimize our LBBAs on COCO-60 dataset. For a fair comparison, we evaluate both two methods with the same COCO-60 dataset (containing 98K images) as the auxiliary dataset. 
When training on our COCO-60 dataset (only removing annotations of VOC classes in COCO dataset) in iteration 0, performance of Zhong \etal drops to $\sim$ 45\% mAP on PASCAL VOC 2007 \textit{test} set (shown in Table~\ref{table:wsod_transfer}). A possible reason is that \emph{the regions with the annotation removed are treated as background in OCUD, which will reduce the recall rate for COCO-60-full}. Compared to Zhong \etal, our LBBA-boosted WSOD is much more stable with data with noise (see Table~\ref{table:per-cls-voc07} for quantitative results). 

In conclusion, our method is different from Zhong \etal, but can be compatible with each other. We believe that the detection performance of WSOD can be better when we apply the method of Zhong \etal into our proposed method.


\end{document}